\documentclass[10pt,twocolumn,letterpaper]{article}

 \usepackage[pagenumbers]{cvpr} % To force page numbers, e.g. for an arXiv version

\definecolor{cvprblue}{rgb}{0.21,0.49,0.74}
\usepackage[pagebackref,breaklinks,colorlinks,allcolors=cvprblue]{hyperref}
\usepackage{booktabs}
\usepackage{tabularx}
\usepackage{array}
\usepackage{makecell}
\usepackage{multirow}
\usepackage{placeins}   
\usepackage{dblfloatfix}
\usepackage{colortbl}
\usepackage[table]{xcolor}
\definecolor{TopOne}{RGB}{142,188,141}   
\definecolor{TopTwo}{RGB}{191,214,186}   
\definecolor{TopThr}{RGB}{247,241,174}  
\newcommand{\best}[1]{\cellcolor{TopOne}\textbf{#1}}
\newcommand{\second}[1]{\cellcolor{TopTwo}#1}
\newcommand{\third}[1]{\cellcolor{TopThr}#1}
\newcommand{\sq}[1]{\textcolor{#1}{\rule{1.1ex}{1.1ex}}} 
\usepackage{siunitx}
\usepackage{threeparttable}
\usepackage{graphicx}
\usepackage{pgfplots} \pgfplotsset{compat=1.18}
\usepackage{listings}
\usepackage{blkarray}   % for blockarray matrix in Eq. (26) example
\usepackage{caption}    % allows \captionsetup{type=listing} when needed
\usepackage{booktabs,makecell,tabularx,threeparttable,placeins,adjustbox,listings,xcolor}
\usepackage{etoolbox} 

\lstdefinestyle{appcode}{
  language=Python,
  basicstyle=\ttfamily\footnotesize,
  numbers=left,
  numbersep=6pt,
  xleftmargin=10pt,
  frame=single,
  framerule=0.3pt,
  rulecolor=\color{black!20},
  keywordstyle=\color{blue!60!black},
  commentstyle=\color{black!55},
  stringstyle=\color{green!40!black},
  breaklines=true,
  tabsize=2
}

\def\paperID{} % *** Enter the Paper ID here
\def\confName{CVPR}
\def\confYear{2026}

\title{Adaptive Multi-Scale Integration Unlocks Robust Cell Annotation in Histopathology Images}

\author{
Yinuo Xu$^{1,3}$ \quad
Yan Cui$^{2,3}$ \quad
Mingyao Li$^{4,*}$\quad
Zhi Huang$^{3,4,*}$ \\
$^{1}$Department of Computer and Information Science, University of Pennsylvania \\
$^{2}$Department of Bioengineering, University of Pennsylvania \\
$^{3}$Department of Pathology and Laboratory Medicine,
University of Pennsylvania \\
$^{4}$Department of Biostatistics, Epidemiology and Informatics,
University of Pennsylvania \\
{\tt\small xuyinuo@cis.upenn.edu, yan12@seas.upenn.edu,}\\
{\tt\small mingyao@pennmedicine.upenn.edu, zhi.huang@pennmedicine.upenn.edu}\\
\small{* To whom the correspondence should be addressed: Zhi Huang, Mingyao Li}
}

\begin{document}
\maketitle
\begin{abstract}
Identifying cell types and subtypes in routine histopathology is fundamental for understanding disease. Existing tile-based models captures nuclear detail but miss the broader tissue context that influences cell identity. Current human annotations are coarse-grained and uneven across studies, making fine-grained, subtype-level classificsation difficult. In this study, we build a marker-guided dataset from Xenium spatial transcriptomics with single-cell–resolution labels for more than two million cells across eight organs and 16 classes to overcome the lack of high-quality annotations.
Leveraging this data resource, we introduce \textbf{NuClass}, a pathologist workflow–inspired framework for \emph{cell-wise} multi-scale integration of nuclear morphology and microenvironmental context. It combines \textbf{Path~local}, which focuses on nuclear morphology from \(224^{2}\)-pixel crops, and \textbf{Path~global}, which models the surrounding \(1024^{2}\)-pixel neighborhood, through a learnable gating module to balance local and global details. An uncertainty-guided objective directs the global path to prioritize regions where the local path is uncertain, and we provide calibrated confidence estimates and Grad-CAM maps for interpretability.
Evaluated on three fully held-out cohorts, NuClass achieves up to \textbf{96\%} F1 for its best-performing class, outperforming strong baselines. Our results demonstrate that multi-scale, uncertainty-aware fusion can bridge the gap between slide-level pathological foundation models and reliable, cell-level phenotype prediction.
\end{abstract}
    
\begin{figure*}[t!]
\centering
\includegraphics[width=\textwidth]{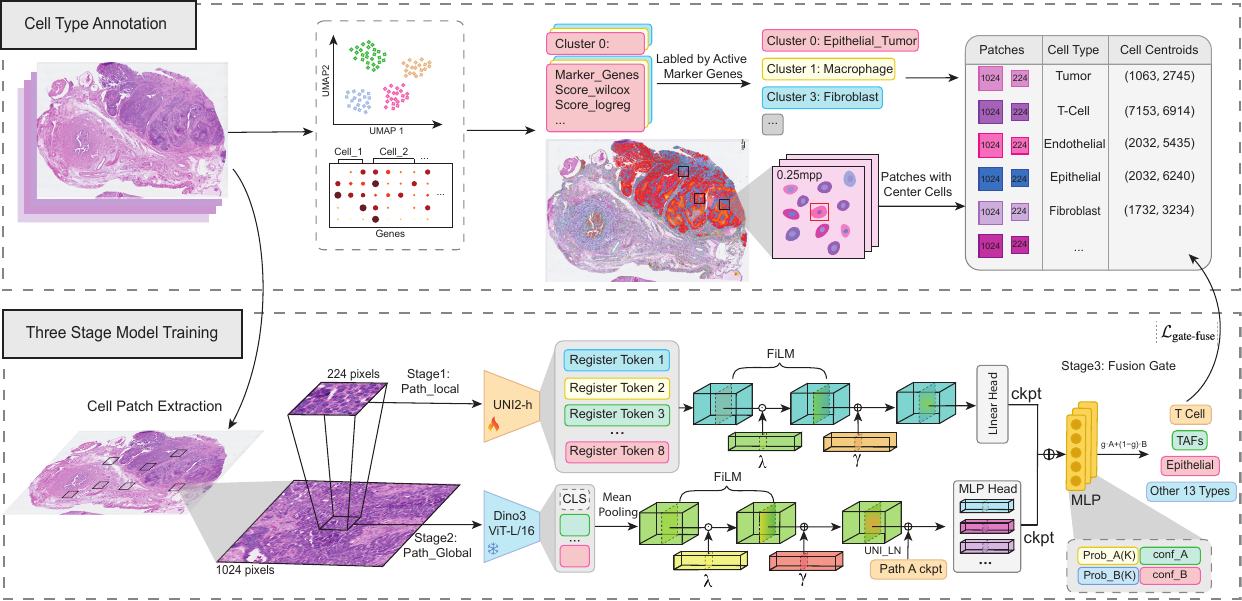}
\caption{\textbf{NuClass: cell-wise multi-scale classification with gating.}
\textbf{(Top)} From whole-slide images (WSIs), we extract paired fields centered on each cell: a nucleus-scale crop ($224^{2}$ at 0.25 mpp) and a co-registered contextual field-of-view (FOV; default $1024^{2}$ at 0.25 mpp). Spatial assays (Xenium-like) provide gene-marker profiles and cell centroids, enabling scalable, subcellular annotations.
\textbf{(Stage 1, Path~local)} A UNI2-h morphology backbone processes the $224^{2}$ crop. A tissue-conditioned FiLM adaptor modulates features before a linear head outputs per-cell probabilities.
\textbf{(Stage 2, Path~global)} A DINOv3 ViT-L/16 encoder ingests the $1024^{2}$ contextual FOV; its representation is concatenated with a Path~local morphology vector to predict complementary probabilities. \textbf{(Stage 3, Fusion~Gate)}  A lightweight \emph{cell-wise gate} receives statistics from both distributions and compact feature projections, then fuses experts in \emph{probability space} $\mathbf p_{\text{mix}}=(1-g)\mathbf p_{local}+g\mathbf p_{global}$. The gate learns \emph{which expert to trust per cell}, not to average logits.
}
\label{fig:overview}
\vspace{-0.5em}
\end{figure*}

\section{Introduction}
\label{sec:intro}

Understanding cell types and their spatial interplay within tissues is central to studying human disease. While emerging technologies such as spatial transcriptomics enable cell type identification through marker genes, hematoxylin and eosin (H\&E)-stained whole-slide images remain the most practical and widely adopted modality in routine clinical workflows, serving as the cornerstone for differential diagnosis and quantitative evaluation and guiding assessment of cancer stage, transplant outcomes, and treatment strategies.

In H\&E images, nuclei segmentation~\cite{stardist} is well studied, but assigning detected nuclei to robust, generalizable cell identities—i.e., cell classification—remains challenging. This largely stems from (1) the lack of large-scale, cross-slice, high-quality, unified training datasets and (2) the difficulty of learning cell type–specific features under heterogeneous morphology across patients, organs, and scanners. Together, these issues reflect two core limitations: \emph{data scarcity} and \emph{model design}.

On the data side, recent studies show that a lack of \emph{high-quality, cell-level annotated training data}, combined with stain/tissue heterogeneity and domain shifts, strongly constrains deep models for cell classification. Human annotations are often coarse and inconsistent and require substantial expert effort, limiting subcellular precision, scalability, and the diversity of labeled cell types.

On the modeling side, three challenges remain. (i) Current cell embedding models are largely unaware of a cell’s neighborhood, even though distinguishing, for example, cancer-associated from normal epithelial cells or macrophages from T cells often requires context beyond the local nuclear image. Multi-scale pipelines for gigapixel slides improve slide-level reasoning~\cite{chen2022hipt}, but aggregation typically occurs at the bag or slide level, leaving the instance-wise balance between morphology and context underexplored. (ii) Models struggle with varying annotation granularities and taxonomies across tissues and datasets (e.g., the same cell may be labeled as ``inflammatory cell,'' ``immune cell,'' or ``lymphocyte''). (iii) There is no large-scale pretrained foundation model across diverse diseases and organs to provide a strong encoder for downstream fine-tuning.

These spatial and semantic nuances across data, annotations, and models remain underexplored, creating an urgent need for a foundation model trained on millions of cells that can harmonize diverse annotations and enable deeper understanding of cell types and tissue organization.

To address data scarcity and model design jointly, we present a workflow from data curation to model development to advance understanding of cellular and nuclear features in H\&E-stained images. On the data side, we build a scalable, \emph{marker-guided} annotation pipeline on Xenium spatial assays that assigns fine-grained cell types with pixel-level coordinates to millions of cells, complementing and refining human supervision~\cite{xeniumtech}. The resulting labels are objective and reproducible across tissues and institutions and provide precise anchors for cross-scale modeling and evaluation (Fig.~\ref{fig:overview}, top).

On the modeling side, we develop \textbf{NuClass}, an adaptive framework for \emph{cell-wise} multi-scale integrated learning. NuClass uses two fixed fields of view: \textbf{Path~local} focuses on nucleus-centered morphology using small crops, whereas \textbf{Path~global} processes the co-registered larger neighborhood to capture tissue architecture and micro-environmental cues. A learnable gate decides, for each cell, how much to rely on local detail versus global context, and training encourages the two paths to specialize and complement rather than duplicate one another.

At scale, NuClass generalizes across organs and datasets for reliable cell-level phenotyping. Trained on more than 2M marker-labeled cells spanning 16 types across 8 organs and evaluated on three fully held-out cohorts (about 0.28M cells), it achieves up to \textbf{96\%} for its best-performing cell class and consistently matches or surpasses strong baselines while producing well-calibrated, interpretable predictions.

\section{Related Work}
\label{sec:formatting}
\newcommand{\parahead}[1]{\noindent\textbf{#1} }

\textbf{Multi-organ multi-class H\&E cell datasets.}
Recent dataset efforts have shifted from single-organ toward multi-organ collections, such as MoNuSAC~\cite{verma2021monusac} (4 tissues) and PanNuke~\cite{gamper2019pannuke} (19 organs/tissues).
However, due to the scarcity of large, well-annotated cell datasets, most cell-classification models remain tailored to specific organs or narrow use-cases, as in HoVer-Net~\cite{graham2019hovernet} and CellViT~\cite{CellViT}.

\textbf{Foundation models for cell classification.}
Foundation models are reshaping computational pathology by learning general-purpose visual or vision--language representations from large, heterogeneous datasets~\cite{chen2024uni,xu2024gigapath,vorontsov2024virchow,lu2024conch,lu2024pathchat,ding2025titan,xiang2025musk}.
These approaches transfer well on patch- and slide-level benchmarks, and vision--language systems extend to retrieval, question answering, and outcome prediction.
Yet most progress still targets slide-level endpoints.
In routine practice, a cell’s identity depends on both its nuclear morphology and the tissue neighborhood in which it resides, but existing systems seldom integrate these two sources of evidence at \emph{per-cell} granularity.
The recent VOLTA algorithm~\cite{nakhli2024volta} leverages over 800k cells across six cancer types to train a self-supervised cell representation model via contrastive learning, but it still relies on cell patches and spatial neighborhoods as unlabeled training signals and does not provide label-consistent, cross-organ cell types.

\parahead{Per-cell representation learning at scale.}
The transition from patch-level to cell-level understanding marks a fundamental step toward morphological precision.
Traditional architectures like HoVer-Net~\cite{graham2019hovernet} and CellViT~\cite{horst2024cellvit} jointly segment and classify cells, yet remain trained on a limited set of annotated organs.
Graph- or structure-aware models~\cite{lou2024senc} extend this by modeling cell neighborhoods and propagating context among adjacent cells to capture micro-environmental cues.
More recently, VOLTA~\cite{nakhli2024volta} proposed environment-aware contrastive learning, jointly encoding nuclear morphology and tissue surroundings to obtain self-supervised embeddings over 0.8M cells; however, these embeddings remain unlabeled and cannot be directly aligned across organs or histologic types.
Beyond isolated cells, cross-organ datasets such as PanNuke~\cite{gamper2019pannuke,gamper2020pannuke_ext}, Lizard~\cite{graham2021lizard}, and NuCLS~\cite{amgad2022nucls} highlight the importance of organ diversity for robust per-cell modeling, yet per-cell recognition still struggles to unify morphology and context across domains.
A scalable foundation model explicitly designed for per-cell representation—bridging nuclear texture, multi-scale tissue context, and cross-organ semantics—therefore remains an open problem in computational pathology.

% ----------------------------- Method -----------------------

\section{Method}
\label{sec:method}

Given a cell–centric dataset with tissue annotations
\(
\mathcal{D}=\{(\mathbf{x}^{224}_i,\mathbf{x}^{ctx}_i,t_i,y_i)\}_{i=1}^N
\),
\(\mathbf{x}^{224}_i\!\in\!\mathbb{R}^{224\times224\times3}\) denotes a local \(224^{2}\) patch,
\(\mathbf{x}^{ctx}_i\!\in\!\mathbb{R}^{H\times W\times3}\) (\(H{=}W{=}1024\) by default) a wider contextual view,
\(t_i\) the tissue type (e.g., liver/lung), and \(y_i\in\{1,\dots,C\}\) the target cell class.
We introduce two complementary branches: \textbf{Path~local} focuses on \emph{cell morphology} with tissue-aware adaptation, while \textbf{Path~global} injects \emph{context} to cover Path~local’s blind spots.
For fusion, we \emph{freeze} the two branches and train a fusion gate to combine predictions in \emph{probability space}, yielding stable gains without modifying either backbone.

\parahead{Notation.}For a \(C\)-class softmax output we write \(\mathbf p\in\Delta^{C-1}\) to mean that
\(\mathbf p=(p_1,\dots,p_C)^\top\in\mathbb R^C\) lies on the probability simplex
\(\Delta^{C-1}=\{\,\mathbf u\in\mathbb R^C:\ u_c\ge 0,\ \sum_c u_c=1\,\}\).
We denote by \(p_{\max}=\max_c p_c\) the largest class probability (“confidence”),
by \(\mathrm H(\mathbf p)=-\sum_{c=1}^C p_c\log p_c\) the Shannon entropy~\cite{shannon1948},
and by \(\mathrm{margin}(\mathbf p)=p_{(1)}-p_{(2)}\) the gap between the largest
and second–largest coordinates (a simple confidence margin).
For an input example \(x_i\), \(\mathbf p_{\text{local}}\) and \(\mathbf p_{\text{global}}\) are the
softmax probability vectors predicted by Path~local and Path~global, respectively;
\(\mathbf p_{\text{mix}}\) denotes the final fused distribution (a convex
combination used at test time).

%----------------------------- Path-local -------------------
\subsection{Path~local: Tissue-aware Cell Morphology Classifier}
\label{sec:method:local}

\parahead{Motivation and task.}In high-throughput pathology, \emph{morphology} is the most direct signal, path~local aims to learn robust, transferable morphology from \(224^{2}\) patches by \emph{adapting to tissue domains}.

\parahead{Architecture.}We adopt a UNI2-h ViT backbone \(\mathcal{F}_{uni}\)~\cite{chen2024uni,dosovitskiy2021vit} to extract a global feature
\(
\mathbf{h}_i=\mathcal{F}_{uni}(\mathbf{x}^{224}_i)\in\mathbb{R}^{1536}
\).
LayerNorm~\cite{ba2016layernorm} and FiLM~\cite{perez2018film} (conditioned on tissue) stylize the feature:
\begin{align}
\mathbf{e}_{t} &= \mathrm{Emb}(t_i)\in\mathbb{R}^{d_t},\\
(\boldsymbol{\gamma}_i,\boldsymbol{\beta}_i) &= \phi(\mathbf{e}_{t})\in\mathbb{R}^{1536}\times\mathbb{R}^{1536},\\
\tilde{\mathbf{h}}_i &= \mathrm{LN}(\mathbf{h}_i)\odot\bigl(1+\boldsymbol{\gamma}_i\bigr)+\boldsymbol{\beta}_i,
\end{align}
where \(\phi(\cdot)\) is a 2-layer MLP and \(\odot\) denotes elementwise product.
A 2-layer MLP head outputs logits and probabilities:
\(
\mathbf{z}_i^{\text{local}}=\mathrm{Head}_{\text{local}}(\tilde{\mathbf{h}}_i)\in\mathbb{R}^{C},\ 
\mathbf{p}_{\text{local}}=\mathrm{softmax}(\mathbf{z}_i^{\text{local}})
\).

\parahead{Loss and schedule.}We use class-balanced cross-entropy with label smoothing~\cite{cui2019classbalanced,muller2019label_smoothing}:
\begin{align}
\mathcal{L}_{\text{local}}
= - w_{y_i}\!\!\sum_{c=1}^{C}\! q_{i,c}\log p_{\text{local},c},\quad
q_{i,c}=
\begin{cases}
1-\epsilon,& c=y_i,\\
\frac{\epsilon}{C-1},& c\neq y_i,
\end{cases}
\end{align}
with class weights from the \emph{Effective Number}:
\(
w_{c}\!\propto\!\frac{1-\beta}{1-\beta^{n_c}}
\) (\(n_c\): samples of class \(c\)).
We adopt a \emph{probe warm-up} followed by full fine-tuning~\cite{kolesnikov2020bit}, and apply layer-wise LR decay to the ViT for stability~\cite{bao2022beit,touvron2022deit3}.

Path~local emphasizes \emph{cell-intrinsic morphology}. Yet \(224^{2}\) local windows can be insufficient under complex micro-environments or cell–cell interactions, motivating Path~global to supply \emph{context}.

% ----------------------------- Path-global -----------------------------
\subsection{Path~global: Context-aware Complementary Branch}
\label{sec:method:global}

\parahead{Motivation and task.}Many cell types become separable only when neighborhood cues (e.g., glands, vessels, immune infiltration) are visible; a \(224^{2}\) crop alone is often insufficient to tell them apart. Path~global \emph{injects wide-field context} while preserving Path~local’s morphology competence, and—crucially—learns \emph{where Path~local is weak} instead of relearning what Path~local already solves.

\parahead{Local-initialized backbone.}Path~global reuses the UNI2-h backbone from Path~local on the \(224^{2}\) crop. We initialize UNI, the tissue-aware FiLM, and the local-style head from Path~local’s checkpoint. For the first \(E_{\text{freeze}}\) epochs these modules are frozen, preventing early drift and ensuring Path~global starts from Path~local’s morphology.

\parahead{Architecture (with tissue FiLM in context).}Given a crop \(\mathbf{x}^{224}_i\) and a contextual view \(\mathbf{x}^{ctx}_i\!\in\!\mathbb{R}^{H\times W\times 3}\), we build a two-stream encoder:
\begin{align}
\mathbf{u}_i &= \mathcal{F}_{uni}(\mathbf{x}^{224}_i) \in \mathbb{R}^{1536}, \\
\mathbf{v}_i &= \mathcal{G}_{ctx}(\mathbf{x}^{ctx}_i) \in \mathbb{R}^{1024}, \quad \text{(DINOv3; token mean)~\cite{simeoni2025dinov3}}
\end{align}
apply tissue-conditioned FiLM to the context stream,
\begin{align}
(\boldsymbol{\gamma}^{\text{ctx}}_i,\boldsymbol{\beta}^{\text{ctx}}_i) &= \phi_{\text{ctx}}(\mathbf{e}_{t}),\\
\tilde{\mathbf{v}}_i &= \mathrm{LN}(\mathbf{v}_i)\odot\bigl(1+\boldsymbol{\gamma}^{\text{ctx}}_i\bigr)+\boldsymbol{\beta}^{\text{ctx}}_i,
\end{align}
then project and concatenate:
\begin{align}
\mathbf{u}'_i &= \mathrm{LN}(\mathbf{u}_i)\mathbf{W}_{u} \in \mathbb{R}^{512}, \nonumber\\[-2pt]
\mathbf{v}'_i &= \mathrm{LN}(\tilde{\mathbf{v}}_i)\mathbf{W}_{v} \in \mathbb{R}^{512}, \nonumber\\[-1pt]
\mathbf{s}_i  &= [\,\mathbf{u}'_i;\, \mathbf{v}'_i\,] \in \mathbb{R}^{1024},\\[2pt]
\mathbf{z}^{\text{global}}_i &= \mathrm{Head}_{\text{global}}(\mathbf{s}_i), \qquad \mathbf{p}^{\text{global}}_i=\mathrm{softmax}(\mathbf{z}^{\text{global}}_i).
\end{align}
\textbf{Why this fusion is efficient.} (i) Both streams are LayerNorm’ed and projected to a common 512-d space, so their scales are comparable; (ii) the 1024-d concatenation lets \(\mathrm{Head}_{\text{global}}\) learn the mixture \emph{in feature space}, which is expressive yet compute-efficient; (iii) empirically this “LN+linear-to-512, then concat” is low-cost and consistently effective.

\parahead{Context resolution and resampling.}Unless specified otherwise, the context field-of-view (FOV) is a \(1024^{2}\) square centered at the target cell at native slide resolution. For compute efficiency we may \emph{resample the same FOV} to
\(r_{\text{ctx}}\!\in\!\{512,768,1024\}\) (default \(r_{\text{ctx}}{=}1024\)) before DINO. Resampling preserves the spatial FOV—no cropping/padding—only the sampling \emph{density} changes. If an integer ratio exists (e.g., \(1024{\rightarrow}512\)) we use stride decimation; otherwise bilinear interpolation. With patch size 16, DINO processes \(S=(r_{\text{ctx}}/16)^2\) tokens and attention scales as \(\mathcal{O}(S^2)\)~\cite{vaswani2017attention}; we therefore employ \emph{micro-batched} forward (e.g., 16–64) to bound memory and compilation latency without sacrificing throughput. 
For cells near slide boundaries, out-of-range regions are mirrored or reflected so that the effective FOV remains \(1024^{2}\) without introducing artificial padding or empty borders.

\parahead{Local-aware weighting (definition).}Let \(\mathbf{z}^{\text{local}}_i\) be Path~local head logits and \(\mathbf{p}^{\text{local}}_i=\mathrm{softmax}(\mathbf{z}^{\text{local}}_i)\).
We up-weight samples for which Path~local is uncertain by
\begin{align}
w_i \;=\; \bigl(1 - p^{\text{local}}_i(y_i)\bigr)^{\gamma}
\label{eq:aaware}
\end{align}
so \(w_i\) increases when Path~local assigns a lower probability to the ground-truth class; this weighting mirrors the focal reweighting pattern~\cite{lin2017focal}.

\parahead{Complementarity supervision (losses).}Our \textbf{default} objective for Path~global is
\begin{align}
\mathcal{L}_{\text{main}}
&=
\mathbb{E}_i\Big[
  w_i \cdot \ell\big(\mathbf{z}^{\text{global}}_i, y_i\big)
\Big],
\label{eq:mainloss}\\[2pt]
\mathcal{L}_{\text{stable}}
&=
\mathbb{E}_i\big[\ell(\mathbf{z}^{\text{global}}_i, y_i)\big],
\label{eq:stableloss}\\[2pt]
\mathcal{L}_{\text{PathGlobal}}
&=
\mathcal{L}_{\text{main}} \;+\; \lambda_{\text{stable}}\,\mathcal{L}_{\text{stable}},
\label{eq:totalloss}
\end{align}
where \(\ell(\mathbf{z},y)=-\sum_{c=1}^{C} w_c\,q_c(y)\,\log \mathrm{softmax}(\mathbf{z})_c\) is class-balanced CE with label smoothing (Effective-Number weights \(w_c\)~\cite{cui2019classbalanced,muller2019label_smoothing}). Intuitively, \(\mathcal{L}_{\text{main}}\) \emph{up-weights} samples where Path~local is unsure via Eq.~\eqref{eq:aaware}, and \(\mathcal{L}_{\text{stable}}\) prevents Path~global from ignoring easy samples entirely. This simple default—\(\boxed{\;\mathcal{L}=\mathcal{L}_{\text{main}}+\lambda_{\text{stable}}\mathcal{L}_{\text{stable}}\;}\)—worked robustly in all experiments.

\parahead{Regularizers: “use context when helpful”.}We apply two light regularizers to avoid collapse to the crop stream: (i) \textbf{Context dropout} (\(\mathbf{v}'_i\!\leftarrow\!\mathbf{0}\) with prob. \(\rho\)), encouraging Path~global to use context only when present; (ii) \textbf{Context warm-up}, concatenating \([\,\mathbf{u}'_i;\,\alpha\,\mathbf{v}'_i\,]\) with \(\alpha\) ramped from \(\alpha_0\) to \(1\) over the first \(E_\alpha\) epochs. These follow dropout/stochastic depth and curriculum learning~\cite{srivastava2014dropout,huang2016stochasticdepth,bengio2009curriculum}.

\parahead{Why it yields complementarity.}Eq.~\eqref{eq:aaware} shifts gradient mass to the \emph{Path~local-hard} region (small \(p^{\text{local}}_i(y_i)\)), while Eq.~\eqref{eq:stableloss} keeps Path~global calibrated on easy cases. Starting from a Path~local–initialized (and briefly frozen) morphology backbone avoids early drift; warm-up/dropout ensure Path~global exploits context only when it helps. Together, Path~global learns \emph{contextual evidence Path~local lacks} rather than duplicating morphology decisions.

% ----------------------------- Fusion -----------------------------
\subsection{Gated Fusion of Path~local and Path~global}
\label{sec:method:fusion}

\parahead{Motivation and formulation.}
Path~local and Path~global exploit different signals—fine nuclear morphology vs.\ wide-field context—and their logits can reside on different scales with distinct uncertainty profiles. Mixing logits is brittle. We therefore fuse in \emph{probability space} via a scalar, per-cell gate \(g_i\!\in\![0,1]\) (mixture-of-experts):
\begin{equation}
\mathbf{p}_{\text{mix},i} \;=\; (1-g_i)\,\mathbf{p}_{\text{local},i} \;+\; g_i\,\mathbf{p}_{\text{global},i},
\label{eq:pmix}
\end{equation}
which is numerically stable and cannot overshoot either path.

\parahead{Gate inputs and architecture (GT-free).}
The gate reads two compact feature projections and GT-free distributional statistics. Let \(\mathbf{f}^{\text{local}}_i\!\in\!\mathbb{R}^{1536}\) be Path~local's pre-head feature and \(\mathbf{s}^{\text{global}}_i\!\in\!\mathbb{R}^{1024}\) be Path~global's fused feature (UNI patch + DINO context). We form
\begin{equation}
\begin{aligned}
\tilde{\mathbf{f}}^{\text{local}}_i &= \mathrm{LN}(\mathbf{f}^{\text{local}}_i)\mathbf{W}^{\text{local}} \in \mathbb{R}^{d_{\text{local}}},\\
\tilde{\mathbf{s}}^{\text{global}}_i &= \mathrm{LN}(\mathbf{s}^{\text{global}}_i)\mathbf{W}^{\text{global}} \in \mathbb{R}^{d_{\text{global}}}.
\end{aligned}
\end{equation}
We then assemble a statistic vector
\begin{equation}
\begin{aligned}
r_i &= 
[\,p^{\text{local}}_{\max},\,p^{\text{global}}_{\max},\,
\bar{\mathrm H}(\mathbf p_{\text{local}}),\,\bar{\mathrm H}(\mathbf p_{\text{global}}),\\
&\quad\mathrm{margin}(\mathbf p_{\text{local}}),\,\mathrm{margin}(\mathbf p_{\text{global}}),\,
\rho_i,\,|\rho_i|\,],
\end{aligned}
\label{eq:gate_stats}
\end{equation}
where \(p^{\cdot}_{\max}=\max_c p^{\cdot}_c\) is top‑1 confidence, \(\bar{\mathrm H}(\cdot)\) is entropy normalized by \(\log C\), \(\mathrm{margin}(\cdot)=p_{(1)}-p_{(2)}\), and 
\[
\rho_i \;\triangleq\; p^{\text{global}}_{\max}-p^{\text{local}}_{\max}
\]
captures GT-free confidence disagreement between the two paths. A two-layer MLP \(\psi(\cdot)\) outputs
\begin{equation}
g_i=\sigma\!\big(\psi\big([\tilde{\mathbf{f}}^{\text{local}}_i;\tilde{\mathbf{s}}^{\text{global}}_i;r_i]\big)\big).
\label{eq:gate}
\end{equation}

\parahead{Supervision: ``choose whom to trust'' (training-only use of GT).}
To supervise the gate, we define predicted labels
\(
\hat y_{\text{local}}=\arg\max_c p^{\text{local}}_{i,c},\;
\hat y_{\text{global}}=\arg\max_c p^{\text{global}}_{i,c}
\)
and a \emph{soft} preference using the ground-truth class:
\[
\begin{aligned}
\Delta^{(y)}_i &\;\triangleq\; p^{\text{global}}_i(y_i)-p^{\text{local}}_i(y_i),\\[4pt]
\tilde g_i &=
\begin{cases}
1,& \hat y_{\text{global}}{=}y_i \land \hat y_{\text{local}}{\neq}y_i,\\
0,& \hat y_{\text{local}}{=}y_i \land \hat y_{\text{global}}{\neq}y_i,\\
\sigma(\kappa\,\Delta^{(y)}_i),& \text{otherwise}.
\end{cases}
\end{aligned}
\]

\emph{Crucially, \(\Delta^{(y)}_i\) is used only for supervision and loss weighting during training; it is never part of the gate input \(r_i\) and is not needed at test time.}

\parahead{Conflict-aware weighting (training-only).}
We emphasize samples on which the two paths disagree, and modulate by their confidence gap on the ground-truth class:
\begin{equation}
w_i \;=\; 1 
+ \lambda_{\text{conf}} \cdot 
   \mathbf{1}\!\big[(\hat y_{\text{local}}{=}y_i)\oplus(\hat y_{\text{global}}{=}y_i)\big]
+ \lambda_{\Delta}\,|\Delta^{(y)}_i|^{\gamma_\Delta}.
\label{eq:w}
\end{equation}

\parahead{Objective.}
The training loss combines fused NLL, gate BCE, a consistency alignment, and an entropy penalty:
\begin{align}
\mathcal{L}_{\text{mix}}
&= \mathbb{E}_i\!\big[\, w_i \cdot \big(-\log \mathbf{p}_{\text{mix},i}(y_i)\big)\,\big], \\
\mathcal{L}_{\text{gate}}
&= \mathbb{E}_i\!\big[\, w_i \cdot \mathrm{BCE}\big(g_i,\tilde g_i;\,\mathrm{pos\_w}\big)\,\big], \\
\mathcal{L}_{\text{align}}
&= \mathbb{E}_i\!\big[\, w_i \cdot \mathrm{KL}\big(\mathbf p^{\star}_i \,\|\, \mathbf p_{\text{mix},i}\big)\,\big], \\
\mathcal{L}_{\text{ent}}
&= \mathbb{E}_i\!\big[\, \lambda_{\text{ent}} \cdot |\Delta^{(y)}_i| \cdot \mathrm{H}_{\text{Bern}}(g_i)\,\big],
\end{align}
where \(\mathbf p^{\star}_i=\mathbf p_{\text{global},i}\) if \(\tilde g_i>0.5\) and \(\mathbf p^{\star}_i=\mathbf p_{\text{local},i}\) otherwise, and \(\mathrm{pos\_w}\) is an adaptive positive weight from the batch. The full objective is
\begin{equation}
\mathcal{L}_{\text{gate-fuse}}
\;=\;
\mathcal{L}_{\text{mix}}
+ \mathcal{L}_{\text{gate}}
+ \lambda_{\text{align}} \mathcal{L}_{\text{align}}
+ \mathcal{L}_{\text{ent}}
+ \lambda_{\text{aux}}\,\mathcal{L}_{\text{aux}},
\label{eq:gate_total}
\end{equation}
with a small auxiliary term \(\mathcal{L}_{\text{aux}}\) added only when unfreezing heads/backbones.

\parahead{Schedule and deployment (safe gate; GT-free).}
We first train the gate with Path~local/Path~global frozen (gate-only warm-up), then optionally unfreeze lightweight modules with small LRs. 
At test time we use Eq.~\eqref{eq:pmix}. 
For deployment we provide a \emph{safe gate} calibrated on validation data: compute per-path reliability by \emph{predicted} class, form a baseline chooser by \(p_{\max}\!\times\) reliability, and accept the gate’s choice only when
\[
g_i \;>\; \tau \quad\text{and}\quad |\rho_i| \;>\; \gamma,
\]
with \(\rho_i=p^{\text{global}}_{\max}-p^{\text{local}}_{\max}\). This conservative policy is GT-free and empirically avoids regressions vs.\ the best single path while often improving accuracy.

\begin{figure}[t]
  \centering
  \includegraphics[width=\linewidth]{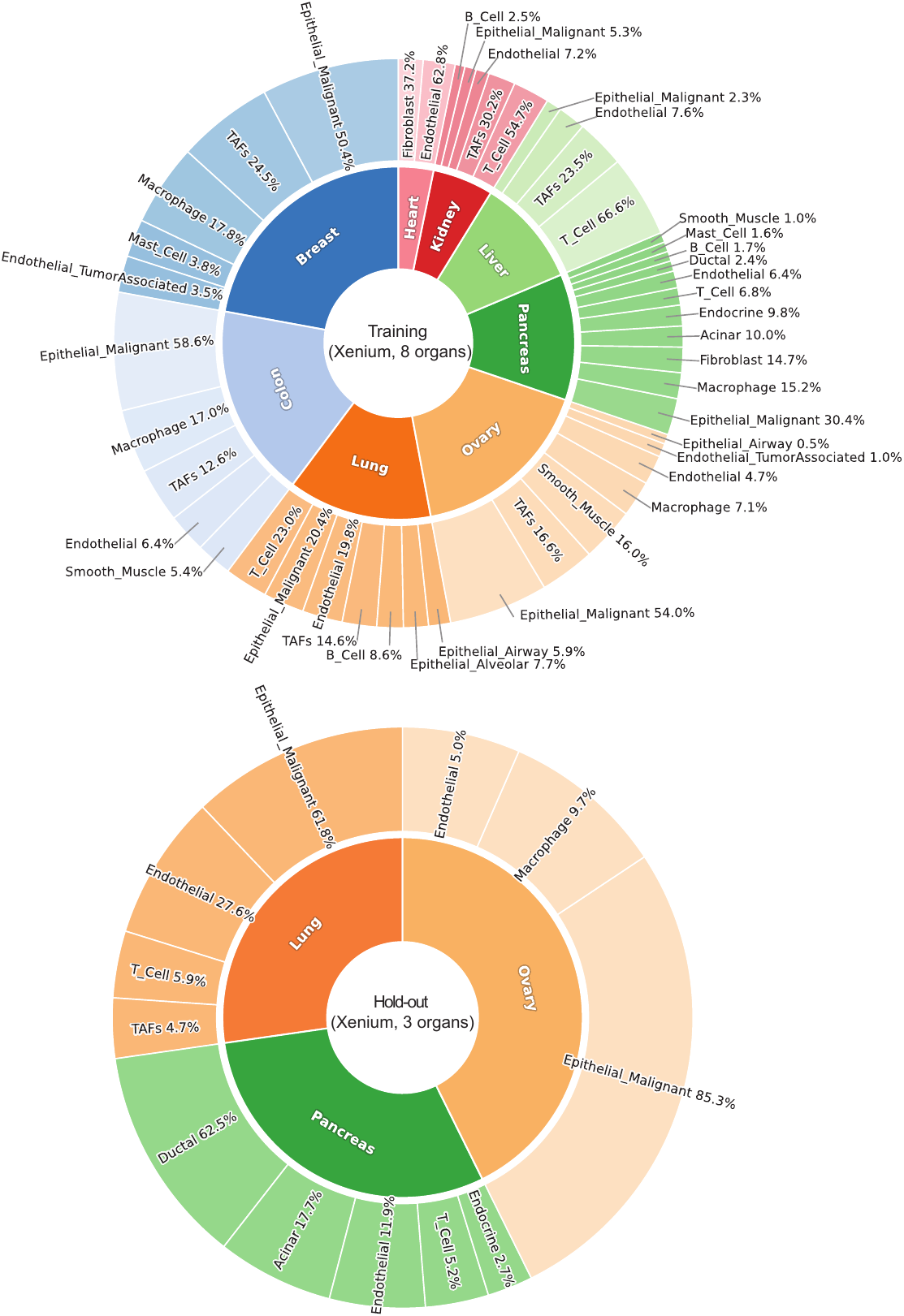}  
  \vspace{-6pt}
  \caption{\textbf{Organ and cohort wise composition.} Donut plots of the training set (8 organs) and the hold-out testing cohorts (pancreas, ovary, lung). The figure illustrates the distribution shifts between training and evaluation cohorts.}
  \label{fig:xenium-donuts}
  \vspace{-4pt}
\end{figure}

\begin{table}[t]
  \centering
  \small
  \setlength{\tabcolsep}{5pt}
  \begin{tabular}{lccc}
    \toprule
    Metric & FiLM & Concat & $\Delta$(F--C) \\
    \midrule
    ACC & 0.8493 & 0.4845 & +0.3648 \\
    F1-macro & 0.5929 & 0.4076 & +0.1853 \\
    Silhouette $\uparrow$ & 0.1480 & 0.0182 & +0.1298 \\
    CH $\uparrow$ & 236.8 & 200.6 & +36.2 \\
    DB $\downarrow$ & 4.26 & 5.56 & –1.30 \\
    ECE $\downarrow$ & 0.061 & 0.213 & –0.152 \\
    \bottomrule
  \end{tabular}
  \vspace{-4pt}
  \caption{FiLM improves feature separability and probability calibration without changing the training pipeline.
  Silhouette/CH/DB follow the classic definitions~\cite{rousseeuw1987silhouette,calinski1974ch,davies1979db};
  ECE follows~\cite{guo2017calibration}.}
  \label{tab:film-ablation}
  \vspace{-6pt}
\end{table}

\section{Experiments}
\label{sec:experiments}

\subsection{Experimental Setup}
\textbf{Datasets and splits.}
All methods are trained on the same multi-organ dataset covering breast, colon, heart, kidney, liver, lung, ovary, and pancreas, and evaluated on three disjoint \emph{hold-out} Xenium cohorts (lung, ovary, and pancreas).
Each image originates from a distinct patient, so there is no patient-level overlap between training and testing cohorts, and no images or annotations from any hold-out cohort are seen during training.
To substantiate the gene-marker-based labeling scheme, Appendix lists all cell types together with their top $10$ marker genes, providing an additional justification for our annotation strategy.
Figure~\ref{fig:xenium-donuts} summarizes the organ and cohort composition.

\textbf{Cohort-specific label alignment.}
We standardize training labels to a canonical taxonomy and remove ambiguous categories.
Each cohort uses a fixed of fixed evaluation labels with a deterministic train$\rightarrow$ evaluation mapping.
Let \(p_{\text{train}}=\mathrm{softmax}(\text{logits}_{\text{train}})\) and let \(\mathbf{M}\in\mathbb{R}^{K_{\text{train}}\times K_{\text{eval}}}\) be a sparse mapping matrix; evaluation-space probabilities are
\begin{equation}
  \mathbf{p}_{\text{eval}} \;=\; \mathbf{M}^\top \mathbf{p}_{\text{train}},
  \label{eq:proj}
\end{equation}

where each column of \(\mathbf{M}\) sums probabilities of training classes that map to the same evaluation class.
For each cohort, the evaluation label set is restricted to the intersection between the canonical taxonomy and the cell types present in that cohort, so that all methods are evaluated on the same set of labels.

\textbf{Information parity \& implementation.}
To equalize metadata usage, all \emph{fine-tuned} models (ours and baselines) attach a learned \textbf{32-d organ embedding} to the visual feature and train it jointly with the classifier,\footnote{Unless noted otherwise, we concatenate the organ embedding to the visual feature.}
while DINO baselines use \emph{frozen backbones with a linear probe} and therefore do not use organ embeddings, preserving the standard LP protocol.
Fine-tuned runs share the same optimizer (AdamW), cosine learning-rate decay with warmup, early stopping on validation macro–F1, PyTorch~2.x with BF16 (TF32 on Ampere+), and identical dataloaders, precision settings, batch sizes, and training schedules per resolution~\cite{loshchilov2019adamw,loshchilov2017sgdr,goyal2017imagenet,paszke2019pytorch}.

\subsection{Comparison with SOTA Models}

\paragraph{Baselines and training setup.}
All baselines are instantiated in the unified label space and data pipeline described above.
For \textbf{MUSK}, \textbf{LOKI}, and \textbf{PLIP} we start from the official checkpoints and recommended training recipes from their original papers and public code, adjusting only image resolution, batch size, and training epochs to match the available GPU memory and keep the total number of optimization steps within the same order of magnitude as our method.
The resulting configurations and GPU usage are summarized in Appendix.

\textbf{Our model.}
We evaluate all three stages of our framework on the same hold-out cohorts:
(1) a \textbf{Path~local} expert specialized for small-scale, densely structured morphology at \(224^2\) resolution;
(2) a \textbf{Path~global} expert capturing large-field contextual patterns at \(1024^2\); and
(3) a learned \textbf{Fusion Gate} that adaptively integrates the two paths by weighting logits per input.
Hyperparameters and training details for these three components are provided in Appendix.

\textbf{Fair comparison protocol.}
All models are trained on the \emph{same} multi-organ corpus, follow the \emph{same} taxonomy and fixed train$\to$eval mapping for each cohort, and use the \emph{same} cohort alignment and evaluation pipeline.
To ensure \emph{information parity}, every fine-tuned model—\textbf{MUSK}, \textbf{PLIP}, \textbf{LOKI}, and ours—receives organ/tissue identity as prior knowledge. For \textbf{DINO} baselines, we keep the frozen-backbone linear-probe setting (no organ conditioning) to preserve the standard LP protocol.
Quantitative results on all cohorts are summarized in Table~\ref{tab:perclass_f1}.

% ---------------- table ----------------
\begin{table*}[t!]
\centering
\setlength{\tabcolsep}{3.6pt}
\renewcommand{\arraystretch}{1.05}
\footnotesize
\caption[Per-class F1 on hold-out datasets.]{
Per-class F1 on hold-out datasets. \emph{Legend:} DINO-B/L/S+/16 = DINOv3 ViT-B/L/S(16+); 
Path~local and Path~global are our single-path experts; 
Fusion Gate is our learned gating fusion. 
Best/second/third are marked as \sq{TopOne}/\sq{TopTwo}/\sq{TopThr}.}
\label{tab:perclass_f1}
\begin{tabular}{llccccccccc}
\toprule
\multicolumn{2}{c}{} & \multicolumn{6}{c}{\textbf{SOTA models}} & \multicolumn{3}{c}{\textbf{Ours}} \\
\cmidrule(lr){3-8} \cmidrule(lr){9-11}
& & \multicolumn{3}{c}{\textit{Frozen backbone + LP}} & \multicolumn{3}{c}{\textit{Fully fine-tuned}} & \multicolumn{3}{c}{} \\
\cmidrule(lr){3-5} \cmidrule(lr){6-8}
Organ & Cell type & DINO-B/16 & DINO-L/16 & DINO-S+/16 & LOKI & PLIP & MUSK & Path local & Path global & Fusion Gate \\
\midrule
\multirow{4}{*}{Lung}
 & Epithelial Malignant & 84.62 & 83.40 & 48.87 & 86.34 & 65.62 & 81.35 & 
    \third{93.90} & \second{95.23} & \best{96.42}\textsuperscript{+1.18} \\

 & Endothelial          & 50.18 & 55.66 & 36.13 & 56.04 & 51.08 & 73.17 & 
    \third{82.49} & \second{83.53} & \best{83.76}\textsuperscript{+0.24} \\

 & T Cell               & 9.72 & 9.28 & 11.99 & 25.49 & 21.73 & \third{28.18} & 
    13.23 & \second{43.89} & \best{54.06}\textsuperscript{+10.17} \\

 & CAFs                 & 35.48 & 24.29 & 9.55  & 25.67 & 19.41 & 28.91 &
    \best{47.55}\textsuperscript{+2.69} & \third{43.74} & \second{44.86} \\
\midrule

\multirow{3}{*}{Ovary}
 & Epithelial Malignant & 90.33 & 90.64 & 82.87 & 64.28 & 87.36 & 83.57 &
    \second{94.14} & \third{93.86} & \best{94.17}\textsuperscript{+0.03} \\

 & Macrophage           & 9.20 & 10.33 & 17.44 & 17.69 & 16.70 & \third{30.17} &
    20.78 & \second{34.78} & \best{35.15}\textsuperscript{+0.37} \\

 & Endothelial          & 22.00 & 20.71 & 7.40 & 28.87 & 35.57 & 53.25 &
    \second{75.20} & \third{73.44} & \best{77.96}\textsuperscript{+2.76} \\
\midrule

\multirow{6}{*}{Pancreas}
 & Endothelial          & 9.68 & 6.63 & 8.33  & 15.28 & 19.67 & 22.49 &
    \best{49.38}\textsuperscript{+0.66} & \third{48.44} & \second{48.72} \\

 & T Cell               & 9.53 & 9.59 & 5.41 & 7.26  & 9.01  & 7.51  &
    \third{20.11} & \second{21.36} & \best{22.54}\textsuperscript{+1.18} \\

 & Fibroblast           & 46.53 &45.37  & 60.86 & 57.48 & 13.20 & \third{65.14} &
    \best{71.58}\textsuperscript{+2.94} & 62.87 & \second{68.65} \\

 & Endocrine            & 0.00 & 0.00 & 8.92  & 0.00  & 1.24  & 20.26 &
    \best{87.39}\textsuperscript{+6.54} & \second{80.85} & \third{68.90} \\

 & Ductal               & 4.29 & 0.00  & 0.00  & 0.00  & 1.66  & 4.12  &
    \best{39.24}\textsuperscript{+4.00} & \third{31.33} & \second{35.24} \\

 & Acinar               & 0.00 & 0.00  & 16.46 & 0.00  & 1.01  & 18.97 &
    \second{44.30} & \third{42.65} & \best{44.60}\textsuperscript{+0.30} \\
\bottomrule
\end{tabular}
\end{table*}

\paragraph{Results summary and takeaways.}
Table~\ref{tab:perclass_f1} reports per-class F1 across all hold-out cohorts.
Overall, our \textbf{Path local} and \textbf{Path global} experts remain strong and complementary, and the learned \textbf{Fusion Gate} typically matches or exceeds the better single path.
For example, on \textbf{lung}, Fusion Gate improves malignant epithelium to \textbf{96.42} F1 and raises T cells from 43.89 (global) to \textbf{54.06} (\(+10.17\)).
On \textbf{ovary}, it increases endothelial from 75.20/73.44 (local/global) to \textbf{77.96}.
These gains reflect the benefit of combining fine-grained morphology with wide-field tissue context.

At the same time, \emph{fusion is not uniformly superior to the best single path}.
In \textbf{pancreas}, several morphology-dominant classes are better handled by \textbf{Path~local}: endocrine (87.39 local vs.\ 68.90 fusion), fibroblast (71.58 vs.\ 68.65), and ductal (39.24 vs.\ 35.24); in \textbf{lung} CAFs, local (47.55) also exceeds fusion (44.86).
We therefore (i) report per-cohort macro/micro F1 and confidence intervals in the Appendix, and (ii) deploy a calibrated \emph{safe gate} that defers to the reliability-weighted single path unless the gate is confident and the two paths strongly disagree.
Empirically this policy is no worse than the best single expert while often preserving the fusion gains on context-sensitive classes.

\noindent\textbf{Takeaways.}
(1) Classes requiring architectural or microenvironmental cues (e.g., lymphocytes, ductal/endocrine, endothelial variants) benefit the most from global context and gated fusion.
(2) For classes where nuclear morphology is decisive, the gate learns to favor Path~local; safe gating further guards against over-fusion.
(3) Under strong class imbalance and distribution shift (e.g., pancreas), multi-scale specialization with adaptive fusion remains robust on average, while acknowledging class-wise variability.

\subsection{Ablation Studies}
\begin{figure}[t]
  \centering
  \includegraphics[width=\linewidth]{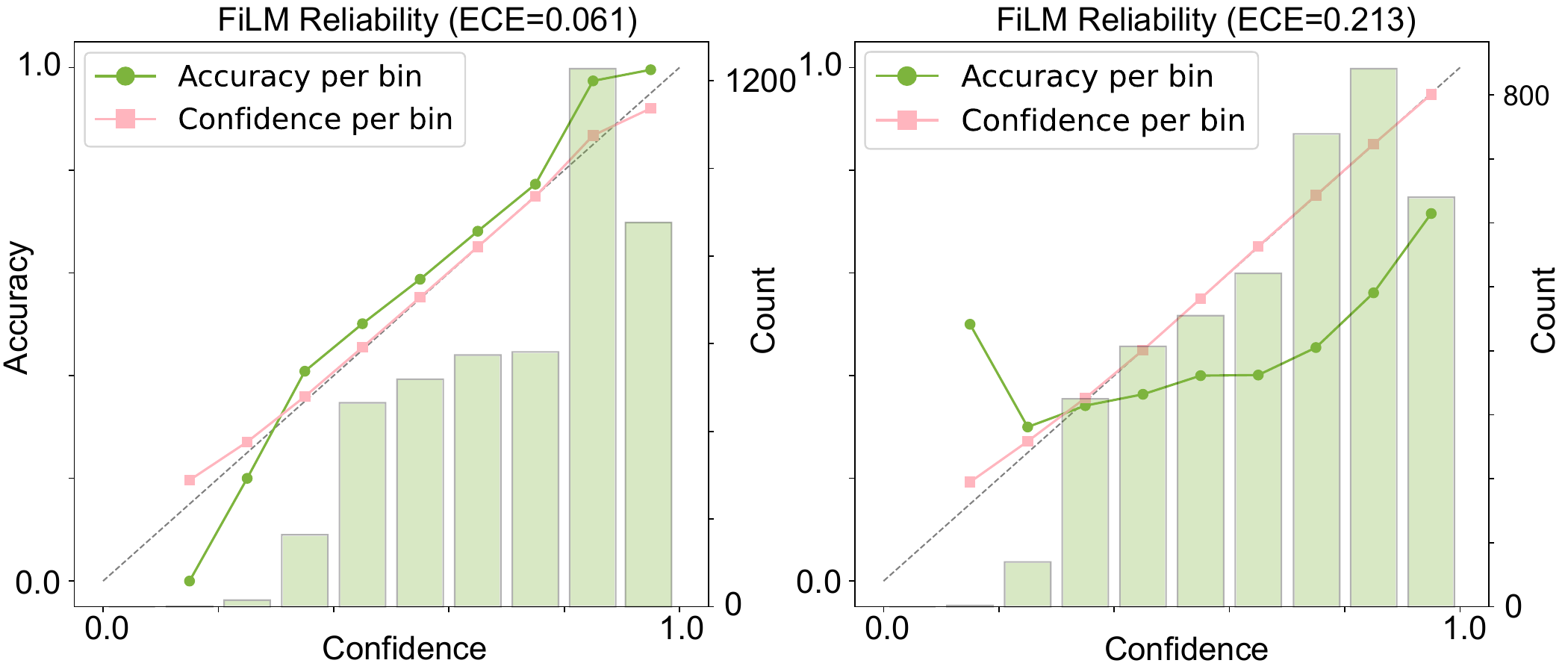}
  \includegraphics[width=\linewidth]{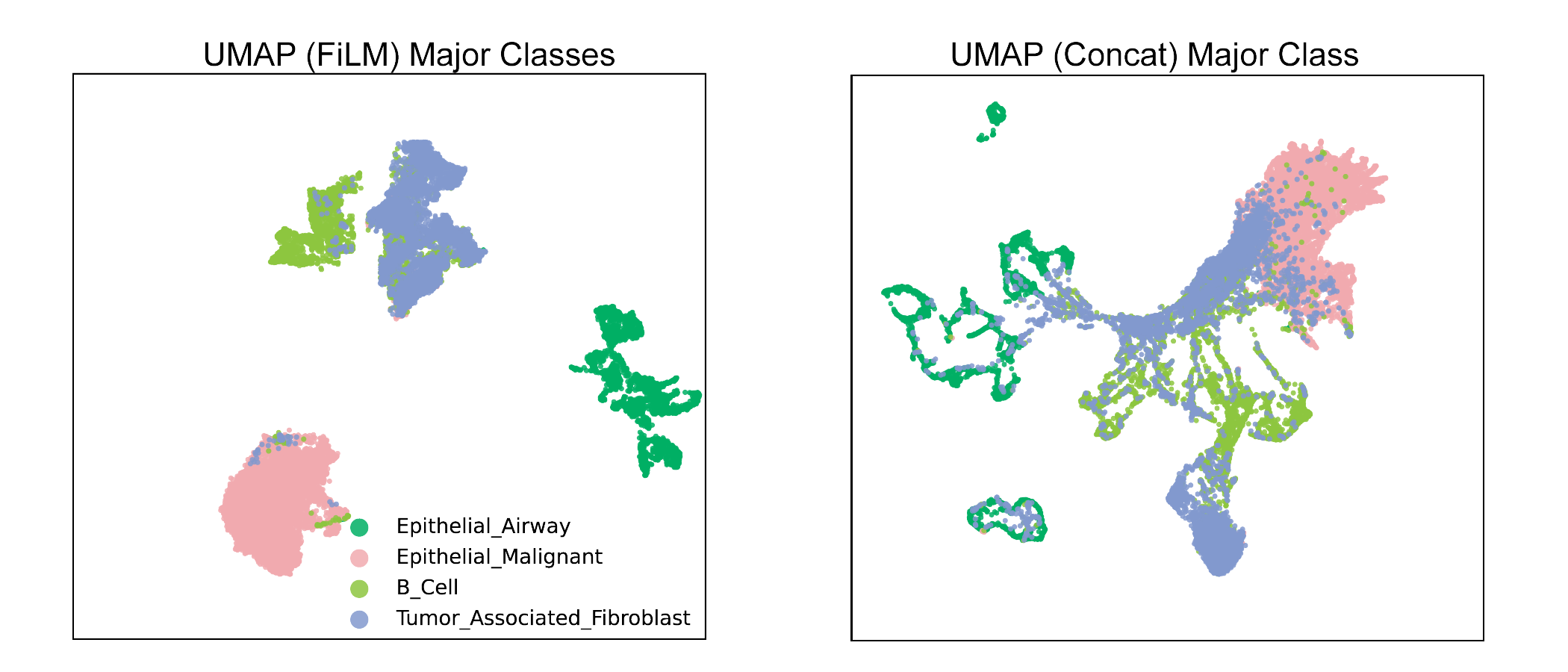}
  \vspace{-6pt}
  \caption{\textbf{Lung subset: reliability and feature geometry.} 
  FiLM (left) vs.\ Concat (right). FiLM reduces ECE~\cite{guo2017calibration} (0.061 vs.\ 0.213) and produces more compact, tissue-aligned clusters.}
  \label{fig:film-lung}
  \vspace{-4pt}
\end{figure}

\textbf{Tissue-aware FiLM in Path~local.}
Replacing simple concatenation with a FiLM modulation between backbone and head improves separability and calibration~\cite{perez2018film}. Given feature \(\mathbf{f}\!\in\!\mathbb{R}^{1536}\) and tissue embedding \(\mathbf{e}_t\!\in\!\mathbb{R}^{64}\), FiLM predicts \((\gamma,\beta)=g_\theta(\mathbf{e}_t)\) and applies \(\tilde{\mathbf{f}}=(1+\gamma)\odot\mathbf{f}+\beta\). On a held-out lung subset, FiLM yields higher accuracy and macro–F1 and lower ECE (Table~\ref{tab:film-ablation}), with tighter clusters and a near-diagonal reliability curve (Fig.~\ref{fig:film-lung}).

\begin{figure}[t]
  \centering
  \includegraphics[width=0.48\textwidth]{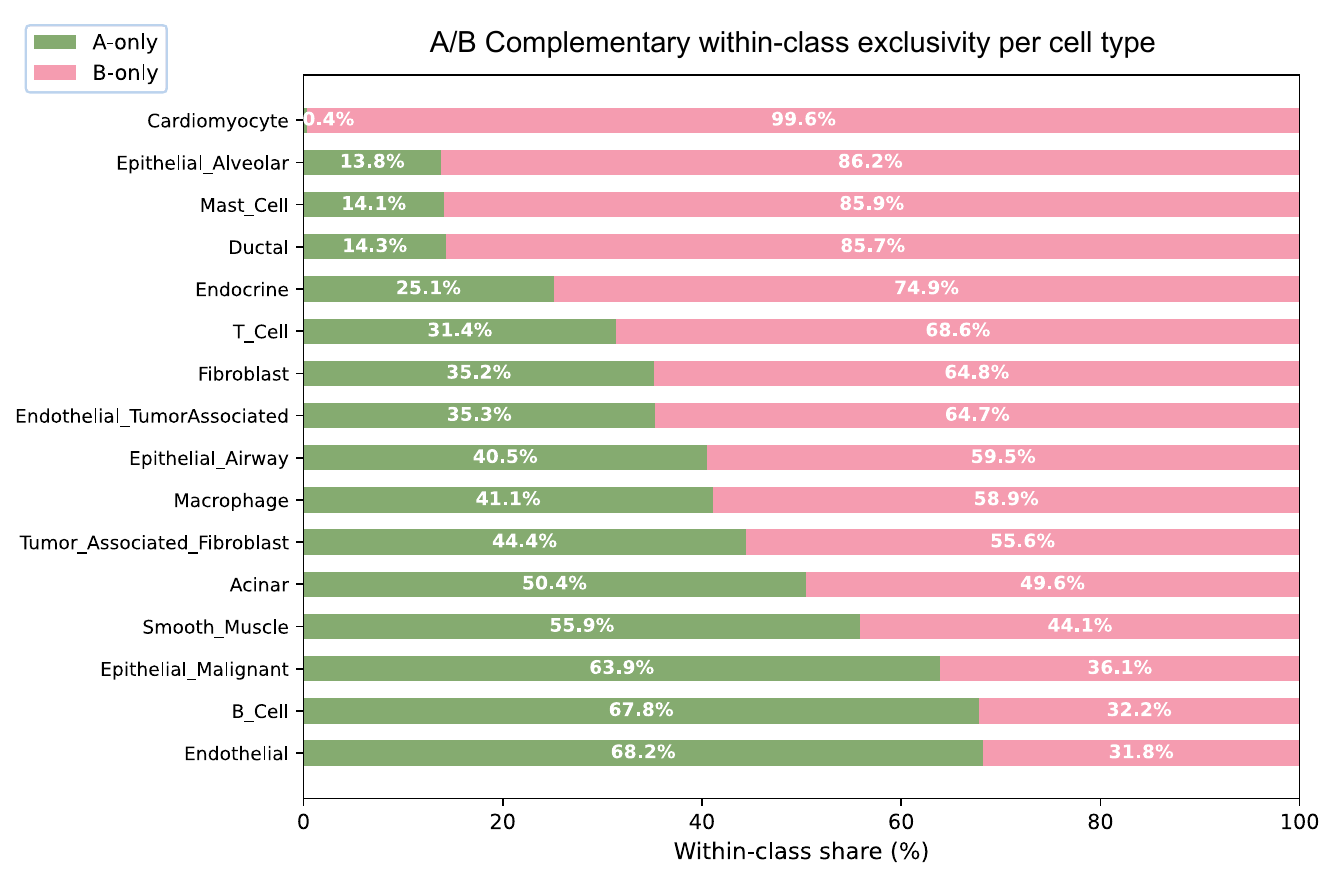}
  \caption{Within-class shares of \emph{local-only} (Path~local correct, Path~global wrong) vs \emph{global-only} (Path~global correct, Path~local wrong). The asymmetry indicates complementary failure modes, motivating gated fusion.}
  \label{fig:ab_complementary}
\end{figure}

\textbf{Complementarity of scales (local/global).}
Evaluating the best checkpoints of \textbf{Path~local} and \textbf{Path~global} on eight validation sets (16 classes total) shows clear disagreement structure: \textbf{Path~global} dominates in 11/16 classes that require broader context (e.g., \emph{cardiomyocyte, ductal, endocrine, T~cell}), while \textbf{Path~local} leads in 4/16 morphology-dense classes (e.g., \emph{B~cell, endothelial, smooth muscle, malignant epithelium}); \emph{acinar} is roughly balanced. An oracle that picks the correct expert per sample yields a positive upper bound, explaining the strong gains of \textbf{Fusion Gate} (Fig.~\ref{fig:ab_complementary}).

\textbf{Fusion-gate loss design.}
The fusion gate is trained with the composite loss
\(
L = L_{\text{mix}} + L_{\text{gate}} + L_{\text{align}} + L_{\text{ent}},
\)
where \(L_{\text{mix}}\) applies cross-entropy to the fused prediction,
\(L_{\text{gate}}\) supervises a soft target that favors the more reliable expert,
\(L_{\text{align}}\) encourages the fused distribution to stay close to the selected expert,
and \(L_{\text{ent}}\) regularizes the gate’s entropy.
We evaluate the contribution of each component through a drop-one ablation on the
eight-organ validation set, using precomputed features and training only the gate
and classification heads. Table~\ref{tab:gate-loss-ablation} reports the detailed results.

\begin{table}[t]
  \centering
  \scriptsize
  \setlength{\tabcolsep}{2.8pt}
  \begin{tabular}{cccccccccc}
    \toprule
    \multirow{2}{*}{$L_{\text{mix}}$} &
    \multirow{2}{*}{$L_{\text{gate}}$} &
    \multirow{2}{*}{$L_{\text{align}}$} &
    \multirow{2}{*}{$L_{\text{ent}}$} &
    \multicolumn{2}{c}{Gate} &
    \multicolumn{2}{c}{Safe} &
    \multicolumn{2}{c}{ECE (Gate / Safe)} \\
    \cmidrule(lr){5-6} \cmidrule(lr){7-8} \cmidrule(lr){9-10}
     & & & &
     Acc & F1 &
     Acc & F1 &
     Gate & Safe \\
    \midrule

    \checkmark & \checkmark & \checkmark & \checkmark
      & 0.8641 & 0.7843
      & 0.8645 & 0.7617
      & 0.0304 & 0.0603 \\

    $\times$ & \checkmark & \checkmark & \checkmark
      & 0.0001 & 0.0000
      & 0.0001 & 0.0000
      & 0.6400 & 0.6515 \\

    \checkmark & $\times$ & \checkmark & \checkmark
      & 0.8663 & 0.7880
      & 0.8647 & 0.7794
      & 0.0221 & 0.0333 \\

    \checkmark & \checkmark & $\times$ & \checkmark
      & 0.8662 & 0.7867
      & 0.8552 & 0.7079
      & 0.0298 & 0.0951 \\

    \checkmark & \checkmark & \checkmark & $\times$
      & 0.8635 & 0.7831
      & 0.8640 & 0.7572
      & 0.0285 & 0.0606 \\

    \bottomrule
  \end{tabular}
  \caption{\textbf{Fusion-gate loss ablation.}
  Columns report performance of the gated prediction (\emph{Gate}) and the Safe fusion~\cite{nygards2016safefusion} output, together with
  calibration error (ECE$\downarrow$).
  Removing $L_{\text{mix}}$ collapses the model, confirming the need for supervision on the
  fused prediction.
  The other terms shape how the gate balances the two experts:
  $L_{\text{align}}$ has a strong impact on Safe-fusion calibration, while $L_{\text{gate}}$
  and $L_{\text{ent}}$ act as mild regularizer that stabilize gating without harming
  accuracy or F1.}
  \label{tab:gate-loss-ablation}
\end{table}

\subsection{Visualization}
\begin{figure}[t]
  \centering
  \includegraphics[width=\linewidth]{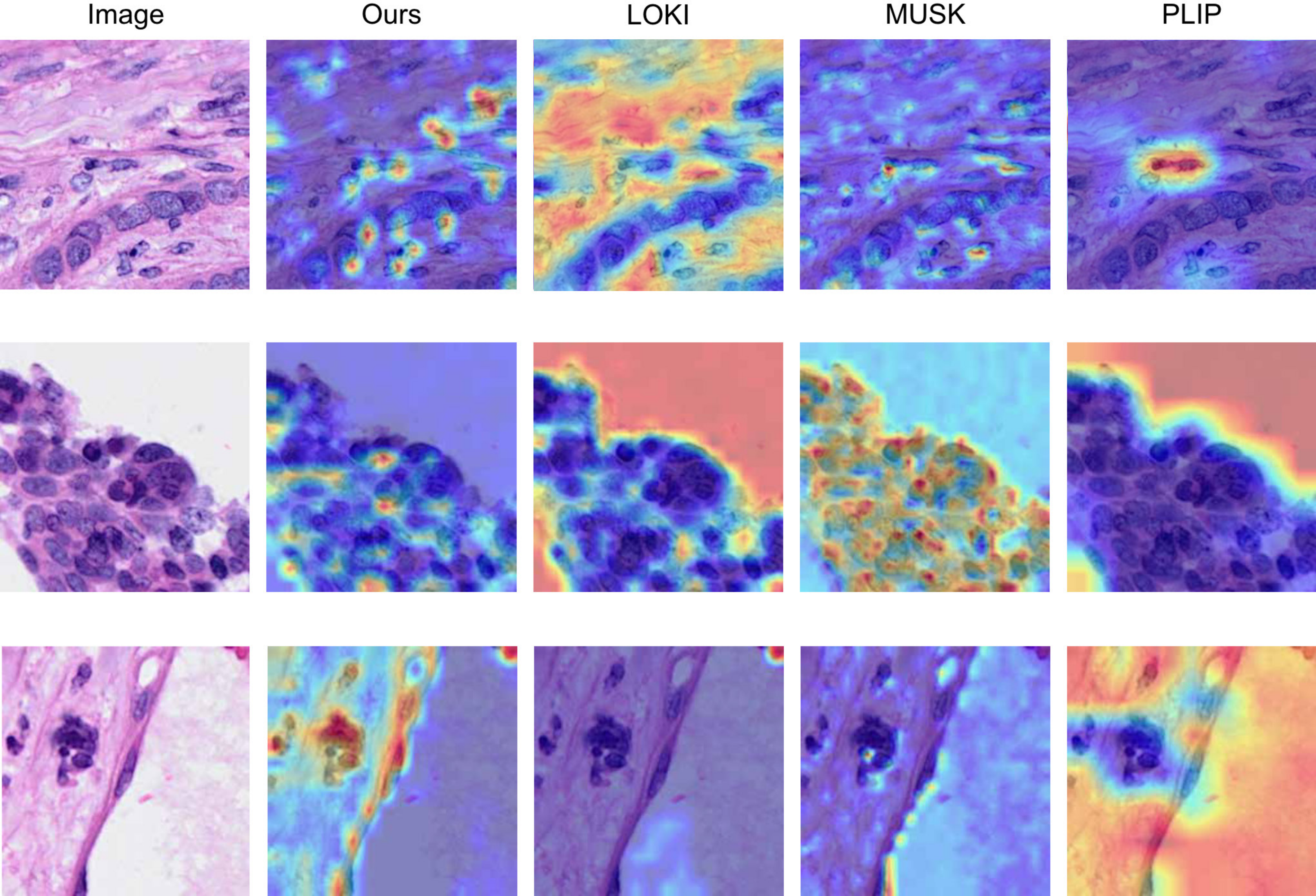}
  \caption{\textbf{Token-level Grad-CAM on ovary patches.}
  Columns: input, \textbf{Path~local}, LOKI, MUSK, PLIP.
  Overlays (last block, $\alpha{=}0.45$) on $224^{2}$ patches.
  Compared with the baselines, \textbf{Path~local} concentrates its
  responses on clusters of nuclei and tissue interfaces, whereas LOKI,
  MUSK, and PLIP often highlight broad texture patterns or even
  background regions, indicating weaker alignment with diagnostic
  structures.}
  \label{fig:ovary-vis}
\end{figure}

To examine \emph{where} different models focus, we visualize token-level
Grad-CAM overlays from the last transformer block on representative
$224^{2}$ ovary patches (Fig.~\ref{fig:ovary-vis})~\cite{selvaraju2017gradcam}.
Across all examples, \textbf{Path~local} produces compact, high-intensity
responses that align with histologically meaningful structures, such as
clusters of nuclei, epithelial–stromal interfaces, and sharp tissue borders.
In contrast, LOKI, MUSK, and PLIP often yield more diffuse or fragmented
activations that primarily follow coarse color/texture patterns and frequently
extend into stromal or background regions with limited diagnostic value.
This tighter spatial correspondence between \textbf{Path~local} attention
and nuclear morphology/boundaries is consistent with its stronger per-class F1 performance. More examples can be found at Appendix.

\section{Conclusion and Discussion}

We presented \textbf{NuClass}, a cell-wise multi-scale framework that treats nuclear morphology and tissue context as independent experts and fuses them through a GT-free, probability-space gate. This design cleanly separates expert specialization from fusion, offers stable training via freeze–then–optional-unfreeze scheduling, and enables a conservative, reliability-calibrated deployment rule. Across three fully held-out cohorts, NuClass consistently matches or exceeds strong frozen and fine-tuned baselines, achieving up to 96\% F1 for its best-performing class and maintaining robust generalization under distribution shift.
Our experiments reveal three central findings:
\begin{itemize}[leftmargin=*]
    \item \textbf{Scale complementarity.} Local morphology and global context excel on different phenotypes; the oracle gains (Fig.~\ref{fig:ab_complementary}) indicate substantial untapped complementarity.
    \item \textbf{Adaptive but selective fusion.} Fusion improves context-sensitive classes, whereas morphology-dominant classes remain better served by the local path, underscoring the need for per-class evaluation.
    \item \textbf{Safe deployment.} A reliability-aware gating policy preserves nearly all fusion benefits while preventing regressions below the strongest single expert.
\end{itemize}

NuClass is directly usable as a plug-in module for existing cell-level pipelines and provides a principled mechanism for deferral when confidence is low. Nonetheless, limitations remain: marker-guided labels inherit Xenium biases; context modeling increases computational cost; and the dataset, while multi-organ, still lacks broader staining, scanner, and population diversity.

Future work will expand NuClass toward pan-cancer coverage, refine taxonomies to finer subtypes and cellular states, extend to end-to-end H\&E-only pipelines with automatic nucleus detection, and explore adaptive context fields, efficient distillation, and stronger uncertainty calibration for clinical deployment. We also aim to incorporate active learning and uncertainty-guided sampling to more efficiently expand the training corpus with high-value annotations and reduce reliance on large-scale marker-based labeling.

{
    \small
    \bibliographystyle{ieeenat_fullname}
    \bibliography{main}
}

% Supplementary (revised, self-consistent with main Eq.(25))
% ==============================

% --- floats & captions ---
\captionsetup{font=small,labelfont=bf}
\newcommand{\tablesmall}{\renewcommand{\arraystretch}{1.08}\setlength{\tabcolsep}{4pt}\small}
\newcolumntype{Y}{>{\raggedright\arraybackslash}X}

% --- appendix counters ---
\makeatletter
\newcommand{\beginappendixcounters}{%
  \setcounter{table}{0}\setcounter{figure}{0}\setcounter{equation}{0}\setcounter{lstlisting}{0}%
  \renewcommand{\thetable}{A\arabic{table}}%
  \renewcommand{\thefigure}{A\arabic{figure}}%
  \renewcommand{\theequation}{A\arabic{equation}}%
  \renewcommand{\thelstlisting}{A\arabic{lstlisting}}%
}
\newcommand{\beginsupplement}{%
  \setcounter{table}{0}\setcounter{figure}{0}\setcounter{equation}{0}\setcounter{lstlisting}{0}%
  \renewcommand{\thetable}{S\arabic{table}}%
  \renewcommand{\thefigure}{S\arabic{figure}}%
  \renewcommand{\theequation}{S\arabic{equation}}%
  \renewcommand{\thelstlisting}{S\arabic{lstlisting}}%
}
\makeatother

% --- code style ---
\lstdefinestyle{appcode}{
  language=Python,basicstyle=\ttfamily\footnotesize,numbers=left,numbersep=6pt,
  frame=single,framerule=0.3pt,rulecolor=\color{black!20},
  keywordstyle=\color{blue!60!black},commentstyle=\color{black!55},
  stringstyle=\color{green!40!black},breaklines=true,tabsize=2
}

\onecolumn
\appendix
\beginappendixcounters

\begin{center}
{\LARGE \textbf{Adaptive Multi-Scale Integration Unlocks Robust Cell Annotation\\
in Histopathology Images}}\\[0.6em]
{\Large \textbf{Supplementary Material}}
\end{center}

% ===============================
% Appendix Overview
% ===============================
\paragraph{Overview of the Supplementary Material}

\begin{figure*}[h]
\centering
\includegraphics[width=\textwidth]{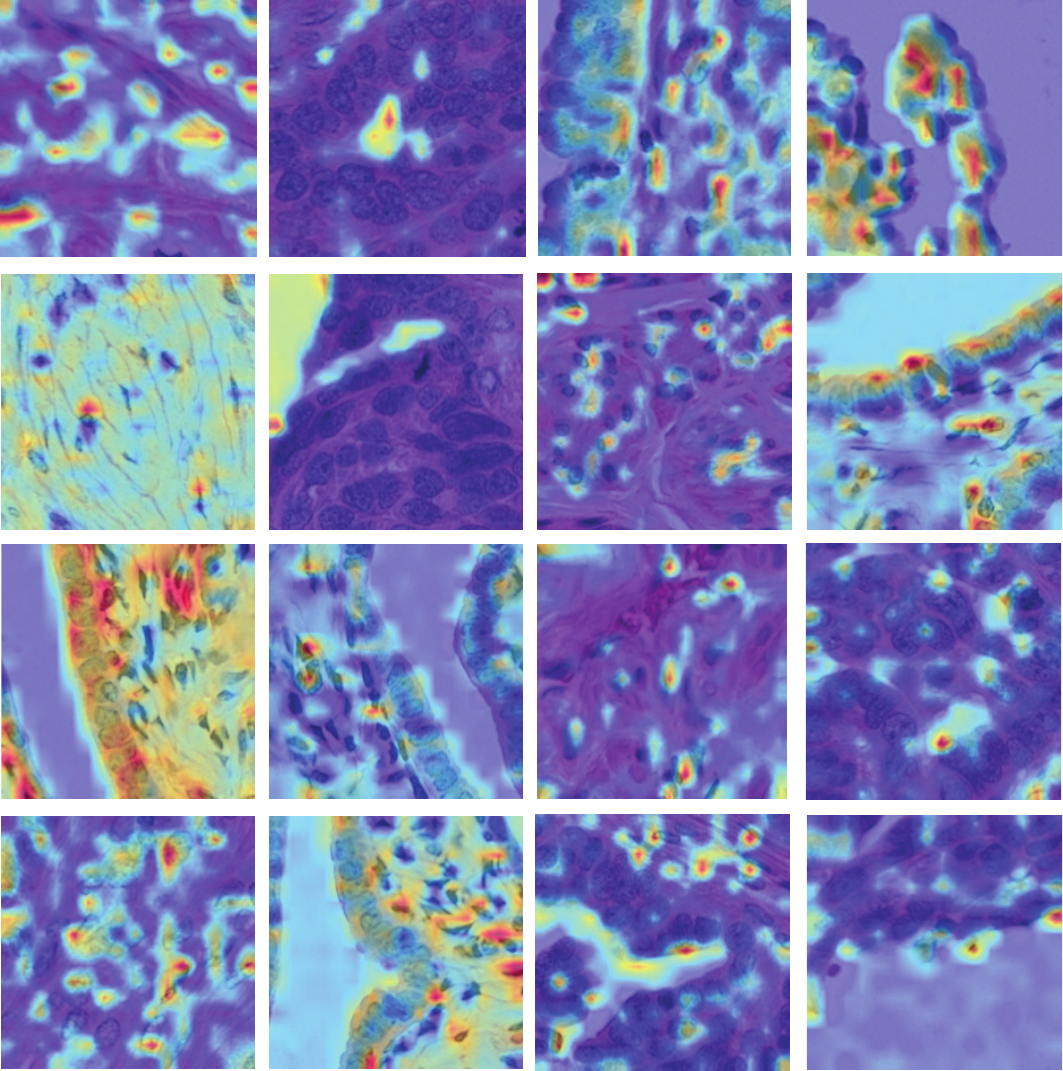}
\caption{\textbf{Additional Grad-CAM visualizations (Path-local expert).}}
\label{fig:appendix_gradcam_pathlocal}
\end{figure*}

\begin{figure*}[h]
\centering
\includegraphics[width=\textwidth]{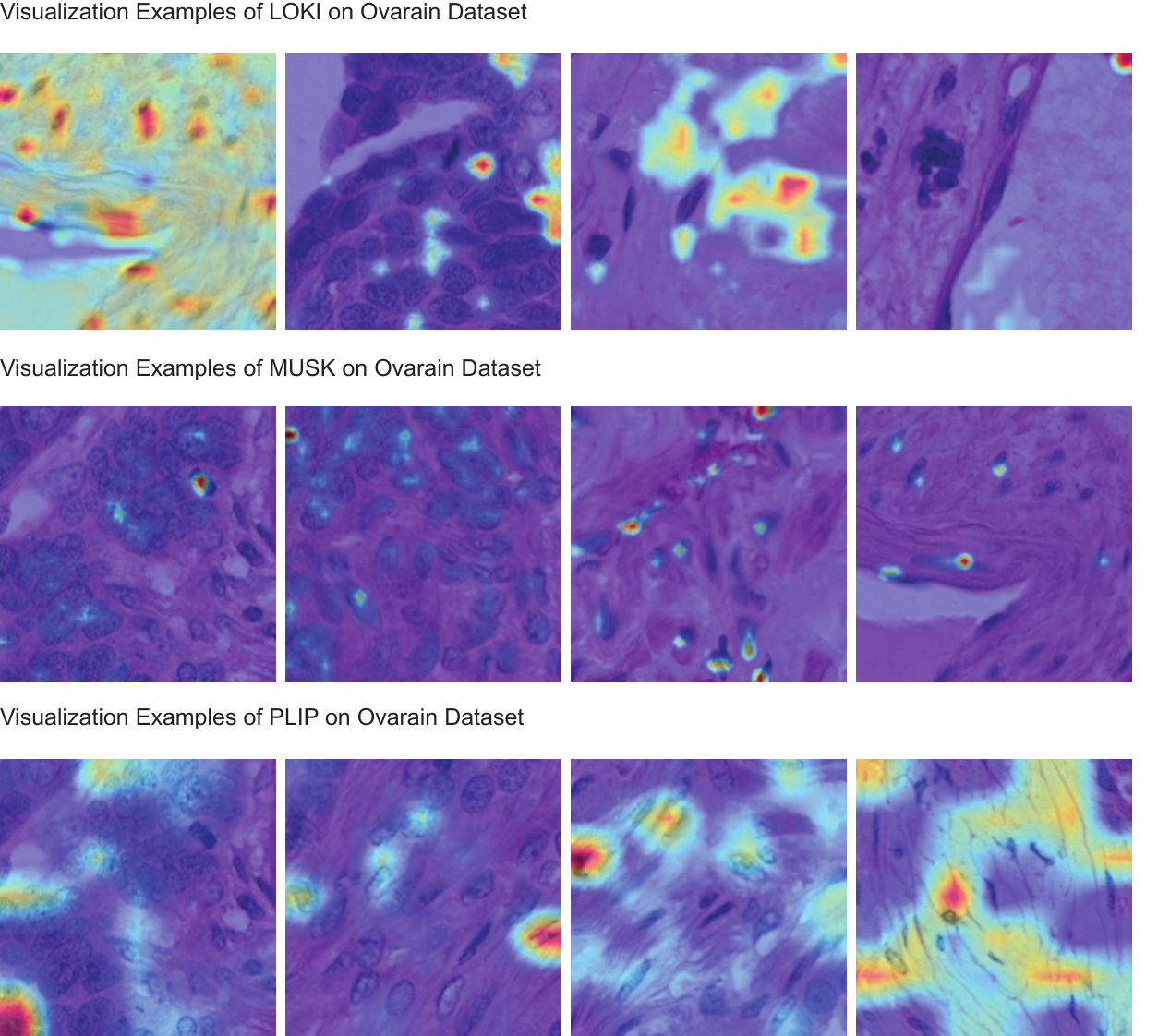}
\caption{Additional Grad-CAM visualizations for LOKI, MUSK, and PLIP (Ovary).}
\label{fig:appendix_gradcam_baseline}
\end{figure*}

\begin{itemize}
  \item \textbf{Baselines: provenance and rationale} (Sec.~\ref{sec:baseline}; Tab.~\ref{tab:a1_baselines}):\\
  we describe how DINOv3 variants (frozen linear probe), PLIP, LOKI, and MUSK are instantiated under a unified protocol, including backbone choices, training protocol (frozen features vs.\ fine-tuning), and the rationale for each baseline family.

  \item \textbf{Full training and inference hyperparameters} (Sec.~\ref{sec:hyperparameters}; Tabs.~\ref{tab:a2_local}--\ref{tab:a2_gate}):\\
  we provide complete optimization settings for the path-local expert, path-global expert, and Fusion Gate, covering optimizers, learning rates, layer-wise decay, warm-up schedules, batch size / precision, data augmentation, tissue conditioning (FiLM), loss formulations, and early stopping. We also describe the organ/tissue prior and the ``unknown organ'' ablation switch used at inference time.

  \item \textbf{Safe Gate calibration and deployment} (Sec.~\ref{sec:hyperparameters}, following Tabs.~\ref{tab:a2_local}--\ref{tab:a2_gate}):\\
  we detail the counting of experts and the Safe Gate training configuration, and explain the validation-only procedure used to select conservative operating points for the gating policy at test time.

  \item \textbf{Compute budget and complexity} (Tab.~\ref{tab:compute-budget}):\\
  we report parameter counts (trainable vs.\ non-trainable), total parameter size, peak GPU memory usage, wall-clock training time, and data scale for the \emph{local}, \emph{global}, and \emph{gate} variants, together with the explicit counting policy that explains why Gate has $\sim 1.67$B non-trainable parameters.

  \item \textbf{Dataset composition, cohort splits, and marker genes} (Sec.~\ref{sec:dataset}; Tabs.~\ref{tab:a3_train}--\ref{tab:a3_dist}):\\
  we summarize per-cohort cell counts, the number of unique cell types, and the train vs.\ hold-out composition after label rebuilding, and we list representative class-level marker genes used for molecular cell-type annotation.

  \item \textbf{Train$\to$eval projection (unified with main Eq.~(25))} (Sec.~\ref{sec:matrices}; Eq.~\ref{eq:proj_unified}):\\
  we formalize the canonical projection $p_{\text{eval}} = M^\top p_{\text{train}}$, provide a minimal numeric example (including tensor shapes), and give explicit cohort-wise binary projection matrices $M_{\text{lung}}$, $M_{\text{ovary}}$, and $M_{\text{pancreas}}$ that map the canonical training taxonomy to each evaluation taxonomy.

  \item \textbf{Cohort-wise Macro/Micro-F1, accuracy, and local-aware gating} (Sec.~\ref{sec:Macro/Micro}; Sec.~\ref{sec:appendix-aaware-pathB}; Tab.~\ref{tab:pathB-aaware-only}):\\
  we report macro-F1 and micro-F1 (accuracy) with 95\% confidence intervals from non-parametric bootstrap ($B{=}1000$) for each held-out cohort and overall, and analyze how the local-aware gating signal affects the global path in terms of accuracy, macro-F1, calibration (ECE, Brier score), AUROC, and performance on low-confidence / high-entropy subsets.

  \item \textbf{Additional visualizations} (Sec.~\ref{sec:visual}; Figs.~\ref{fig:appendix_gradcam_pathlocal}--\ref{fig:appendix_gradcam_baseline}):\\
  we provide additional Grad-CAM visualizations for the path-local expert and for LOKI, MUSK, and PLIP on representative examples, illustrating qualitative differences in attention patterns across methods.
\end{itemize}

\newpage
% ---------------------------------------------------------------
\section{Baselines: provenance and rationale}
% ---------------------------------------------------------------
\label{sec:baseline}
\begin{table}[h]
\centering
\tablesmall
\begin{adjustbox}{max width=\textwidth}
\begin{tabular}{l l l l l}
\toprule
\textbf{Model} & \textbf{Type} & \textbf{Backbone} & \textbf{Training protocol} & \textbf{Rationale} \\
\midrule
DINOv3-B/16    & SSL ViT (frozen) & ViT-B/16        & Frozen features + linear probe & Canonical LP baseline \\
DINOv3-L/16    & SSL ViT (frozen) & ViT-L/16        & Frozen features + linear probe & Larger-scale LP variant \\
DINOv3-S+/16   & SSL ViT (frozen) & ViT-S+/16       & Frozen features + linear probe & Smaller LP variant \\
PLIP           & VLP (fine-tuned) & ViT-L/14        & Vision tower fine-tuning       & Pathology VLP transfer \\
LOKI           & VLP (fine-tuned) & ViT-L/14-like   & Vision tower fine-tuning       & Visual--omics VLP baseline \\
MUSK           & Vision-only (FT) & p16 transformer & End-to-end fine-tuning         & Strong non-VLP foundation model \\
\bottomrule
\end{tabular}
\end{adjustbox}
\caption{Snapshot of baseline families and protocols used in our experiments. All fine-tuned baselines employ the same 32-d organ embedding, data pipeline, and optimization setup for parity; DINOv3 variants are evaluated with frozen backbones and a learned linear probe.}
\label{tab:a1_baselines}
\end{table}

% ---------------------------------------------------------------
\section{Full training and inference hyperparameters}
% ---------------------------------------------------------------
\label{sec:hyperparameters}

\begin{table}[h]
\centering
\tablesmall
\begin{adjustbox}{max width=\textwidth}
\begin{tabular}{l l}
\toprule
\textbf{Backbone} & UNI2-h (ViT-L/14-like, $224^2$, embed=1536) \\
\midrule
Optimizer / Weight Decay & AdamW / $2{\times}10^{-5}$ \\
LR (UNI / Head) & $3{\times}10^{-6}$ / $3{\times}10^{-4}$ \\
Layer-wise LR decay $\gamma$ & $0.85$ (earlier layers $\to$ lower LR) \\
Scheduler & Linear warm-up to 1.0 (steps=$1500$); no decay \\
Epochs / Freeze & $50$ / freeze backbone for the first epoch \\
Batch (effective) / Precision & $512{\times}2{=}1024$ / BF16-mixed \\
Data augmentation & H/V flip, $\pm 15^\circ$ rotation, color jitter, affine, normalization \\
Tissue conditioning & FiLM (embedding dim $64$) \\
Loss & Class-balanced CE (Effective Number, $\beta{=}0.9999$) + label smoothing $\varepsilon{=}0.1$ \\
Early stopping / Validation cadence & patience $4$ / every $0.25$ epoch \\
\bottomrule
\end{tabular}
\end{adjustbox}
\caption{Path-local hyperparameters.}
\label{tab:a2_local}
\end{table}

\begin{table}[h]
\centering
\tablesmall
\begin{adjustbox}{max width=\textwidth}
\begin{tabular}{l l}
\toprule
\textbf{Backbones} & UNI2-h ($224^2$) + DINOv3-L/16 (context) \\
\midrule
Optimizer / Weight Decay & AdamW / $2{\times}10^{-5}$ \\
LR (UNI / Head / DINO) & $2{\times}10^{-6}$ / $3{\times}10^{-4}$ / $1{\times}10^{-6}$ (DINO frozen by default) \\
Layer-wise LR decay $\gamma$ & $0.85$ (UNI) \\
Scheduler & Linear warm-up (steps=$2000$); no decay \\
Epochs / Precision & $40$ / BF16-mixed \\
Batch (effective) & $256{\times}2{=}512$ \\
Context size / micro-batch & $1024$ / $16$ (DINO forward chunking) \\
Fusion $\alpha$ warm-up & $\alpha{:}\;0.25\to 1.0$ (first $2$ epochs) \\
Context dropout & $0.10$ \\
Freeze policy & UNI unfrozen; DINO frozen (optional unfreeze) \\
\midrule
\textbf{Loss} & $w=(1-p_A(y))^\gamma,\ \gamma{=}2.0$; $\mathcal{L}=\mathcal{L}_{\text{main}}+0.3\,\mathcal{L}_B$ \\
\bottomrule
\end{tabular}
\end{adjustbox}
\caption{Path-global hyperparameters.}
\label{tab:a2_global}
\end{table}

\begin{table}[h]
\centering
\tablesmall
\begin{adjustbox}{max width=\textwidth}
\begin{tabular}{l l}
\toprule
\textbf{Inputs / Backbones} & A: UNI2-h+FiLM; B: UNI2-h + DINOv3-L context \\
\midrule
Optimizer / Weight Decay & AdamW / $1{\times}10^{-5}$ \\
LRs (gate / heads / backbone) & $8{\times}10^{-4}$ / $2{\times}10^{-4}$ / $5{\times}10^{-5}$ \\
Scheduler & Linear warm-up (steps=$1000$) \\
Epochs / Batch (effective) & $4$ / $12{\times}32{=}384$ \\
Context size / micro-batch & $384$ / $16$ (context resized on CPU) \\
Gate warm-up & first $0.5$ epoch gate-only, then optional unfreeze \\
Gate MLP & projA(1536$\to$128)+projB(1024$\to$128)$\to$MLP(256,64,1); dropout $0.1$ (gate), $0.2$ (head) \\
Losses & $\mathcal{L}_{\text{mix}}{+}\mathcal{L}_{\text{gate}}{+}\lambda_{\text{align}}\mathcal{L}_{\text{align}}{+}\mathcal{L}_{\text{ent}}{+}\lambda_{\text{aux}}\mathcal{L}_{\text{aux}}$ \\
Weights & $\lambda_{\text{align}}{=}0.1,\ \lambda_{\text{ent}}{=}0.05,\ \lambda_{\text{aux}}{=}0.05$; conflict boost $=2.0$ \\
\bottomrule
\end{tabular}
\end{adjustbox}
\caption{Fusion Gate hyperparameters.}
\label{tab:a2_gate}
\end{table}

\paragraph{Organ/tissue prior and ablation note.}
In routine surgical pathology, organ/tissue identity is recorded at accession and remains available during H\&E processing; we therefore treat it as known metadata and encode it via a 32-d embedding for all fine-tuned models (``information parity''). To simulate ``unknown organ'' without retraining, our released inference code provides a switch that replaces the organ embedding with a learned \texttt{[UNK]} token (or a zero-vector through FiLM), allowing users to assess sensitivity to this prior.

\begin{table}[t]
\centering
\tablesmall
\caption{Compute and memory budget of the \textbf{local}, \textbf{global}, and \textbf{gate} variants. Parameter counts follow PyTorch Lightning summaries; memory numbers are peak usage on a single H200 (141\,GB).}
\label{tab:compute-budget}
\vspace{0.4em}

\begin{adjustbox}{max width=\textwidth}
\begin{tabular}{lcccc}
\toprule
\textbf{Setup} & \textbf{Trainable params} & \textbf{Non-trainable params} & \textbf{Total params} & \textbf{Param size} \\
\midrule
local (baseline) & 683M & 0 & 683M & 2.73\,GB \\
global (ours) & 686M & 303M & 989M & 3.96\,GB \\
gate (ours) & $\approx 0.42$M & $\approx 1.67$B & $\approx 1.67$B & $\approx 6.7$\,GB \\
\bottomrule
\end{tabular}
\end{adjustbox}

\vspace{0.6em}

\begin{adjustbox}{max width=\textwidth}
\begin{tabular}{lcccc}
\toprule
\textbf{Setup} & \textbf{GPU(s)} & \textbf{Peak memory (of 141\,GB)} & \textbf{Wall-clock time} & \textbf{Data scale} \\
\midrule
local (baseline) & 1\,$\times$\,H200 & 15\% (\,$\approx$21\,GB) & $\sim$5\,h & 1.60M train / 0.45M val (16 classes) \\
global (ours) & 1\,$\times$\,H200 & 40\% (\,$\approx$56\,GB) & $\sim$22\,h & 1.60M train / 0.45M val (16 classes) \\
gate (ours) & 1\,$\times$\,H200 & 69\% (\,$\approx$97\,GB) & $\sim$5\,h & 1.60M train / 0.45M val (16 classes) \\
\bottomrule
\end{tabular}
\end{adjustbox}
\end{table}

\paragraph{Counting policy for Table~A6 (why Gate shows 1.67B non-trainable).}
Table~A6 reports parameter counts from PyTorch Lightning ``summary'' under the \emph{modules resident in memory} during each stage.
Gate training instantiates \textbf{both} experts end-to-end and freezes them:
Path~A uses UNI2-h, Path~B uses UNI2-h \emph{and} DINOv3-L/16 for context.
Hence the non-trainable pool is
\[
\underbrace{\text{UNI2-h}}_{\text{Path local}}
+\underbrace{\text{UNI2-h}}_{\text{Path global}}
+\underbrace{\text{DINOv3-L/16}}_{\text{Path global}}
\ \approx\ 2\times 683\text{M}+303\text{M}=1{,}669\text{M},
\]
matching the Gate row (non-trainable/total/size). Only the small Gate MLP and heads ($\sim 0.42$M) are trainable.

\newpage
% ---------------------------------------------------------------
\section{Dataset composition and cohort splits}
% ---------------------------------------------------------------
\label{sec:dataset}
\begin{table}[h]
\centering
\tablesmall
\begin{adjustbox}{max width=\textwidth}
\begin{tabular}{l r r}
\toprule
\textbf{Cohort (organ)} & \textbf{Cells} & \textbf{Unique types}\\
\midrule
Ovarian\_wsi (ovary) & 384{,}805 & 7 \\
Pancreas\_wsi (pancreas) & 187{,}522 & 11 \\
Heart-MultiCancerPanel (heart) & 5{,}859 & 2 \\
Kidney-MultiCancerPanel-Cancer (kidney) & 27{,}550 & 5 \\
Liver-MultiCancerPanel-Cancer (liver) & 96{,}716 & 4 \\
Lung-Cancer-v1+5k (lung) & 195{,}133 & 7 \\
Colon\_Preview (colon) & 446{,}933 & 5 \\
Breast\_Entire-Sample-Area-S1 (breast) & 701{,}220 & 5 \\
\midrule
\textbf{Combined} & \textbf{2{,}045{,}738} & \textbf{16}\\
\bottomrule
\end{tabular}
\end{adjustbox}
\caption{Training cohorts after mapping/filtering.}
\label{tab:a3_train}
\end{table}

\begin{table}[h]
\centering
\tablesmall
\begin{adjustbox}{max width=\textwidth}
\begin{tabular}{l r r}
\toprule
\textbf{Cohort (organ)} & \textbf{Cells} & \textbf{Unique types}\\
\midrule
Pancreatic\_Multimodal\_Cell\_Segmentation (pancreas) & 74{,}689 & 6 \\
Ovarian-ImmunoOnco+CustomAddon (ovary) & 150{,}748 & 3 \\
Lung-ImmunoOnco+CustomAddon (lung) & 61{,}490 & 4 \\
\midrule
\textbf{Combined} & \textbf{286{,}927} & \textbf{9}\\
\bottomrule
\end{tabular}
\end{adjustbox}
\caption{Hold-out cohorts.}
\label{tab:a3_holdout}
\end{table}

\begin{table}[h]
\centering
\tablesmall
\begin{adjustbox}{max width=\textwidth}
\begin{tabular}{l l}
\toprule
\textbf{Class} & \textbf{Top 10 marker genes} \\
\midrule
Epithelial\_Malignant        & EPCAM, KRT8, KRT18, KRT19, KRT17, KRT5, MKI67, MCM2, TOP2A, BIRC5 \\
Cancer\_Associated\_Fibroblast & COL1A1, COL1A2, COL3A1, ACTA2, TAGLN, PDGFRB, FAP, PDPN, POSTN, CXCL12 \\
Macrophage                   & CD68, LST1, CSF1R, MS4A7, C1QA, C1QB, C1QC, CD163, MRC1, ITGAM \\
T\_Cell                      & CD3D, CD3E, CD2, CD4, CD8A, TRAC, CD247, LCK, IL7R, CCR7 \\
Endothelial                  & PECAM1, VWF, KDR, CD34, EMCN, FLT1, KLF2, CLDN5, ENG, ESAM \\
Smooth\_Muscle               & ACTA2, TAGLN, MYH11, CNN1, MYL9, DES, TPM2, CALD1, RGS5, MYLK \\
Mast\_Cell                   & KIT, TPSAB1, TPSB2, CPA3, MS4A2, HDC, GATA2, FCER1A, SRGN, SLC18A2 \\
Fibroblast                   & COL1A1, COL1A2, DCN, LUM, PDGFRA, COL3A1, COL5A1, FBLN1, THY1, VIM \\
Endothelial\_CancerAssociated & PECAM1, VWF, KDR, ICAM1, SELE, FLT1, VCAM1, ANGPT2, IL6, CXCL8 \\
B\_Cell                      & MS4A1, CD79A, CD79B, CD74, CD19, CD22, CD37, CD40, BANK1, PAX5 \\
Acinar                       & PRSS1, PRSS3, CPA1, CPB1, CELA3A, CTRB1, CTRC, CEL, CPA2, REG1A \\
Endocrine                    & INS, GCG, SST, PPY, CHGA, IAPP, PCSK1, PCSK2, PAX6, MAFA \\
Epithelial\_Alveolar         & SFTPA1, SFTPA2, SFTPB, SFTPC, SLC34A2, AGER, LPCAT1, ABCA3, KRT8, KRT18 \\
Epithelial\_Airway           & KRT5, KRT8, KRT18, MUC1, SCGB1A1, KRT14, KRT15, KRT17, FOXJ1, EPCAM \\
Ductal                       & KRT7, KRT8, KRT18, KRT19, MUC1, EPCAM, KRT23, SOX9, CFTR, SPP1 \\
epithelial                   & EPCAM, KRT8, KRT18, KRT19, KRT5, KRT14, KRT15, KRT17, CDH1, KRT7 \\
lymphocyte                   & PTPRC, CD3D, CD3E, MS4A1, CD79A, CD4, CD8A, NKG7, GNLY, TRAC \\
plasma                       & MZB1, XBP1, SDC1, SLAMF7, PRDM1, IGKC, IGHG1, JCHAIN, TNFRSF17, CD38 \\
\bottomrule
\end{tabular}
\end{adjustbox}
\caption{Class-level marker genes used for cell type annotation. We list the top 10 marker genes per class here for brevity; in our actual analysis we use the top 50 marker genes.}
\label{tab:a3_dist}
\end{table}

% ---------------------------------------------------------------
\section{Cohort-specific projection matrices}
% ---------------------------------------------------------------
\label{sec:matrices}
\paragraph{Definition and notation (unified).}
Let $C_{\text{train}}$ and $C_{\text{eval}}$ be the numbers of classes in the training and evaluation taxonomies for a given cohort.
We define a fixed binary projection
\[
M \in \{0,1\}^{C_{\text{train}}\times C_{\text{eval}}},\qquad
M_{ij}=1\ \text{iff training class } i\ \text{maps to evaluation class } j.
\]
Given $p_{\text{train}}\in\Delta^{C_{\text{train}}-1}$, the evaluation-space probabilities are
\begin{equation}
\label{eq:proj_unified}
p_{\text{eval}} \;=\; M^{\top} \, p_{\text{train}} \;\in\; \mathbb{R}^{C_{\text{eval}}},
\end{equation}
so $\big(p_{\text{eval}}\big)_{j}=\sum_{i:\,M_{ij}=1}\big(p_{\text{train}}\big)_{i}$.
We rebuild labels so that each retained training class maps to \emph{exactly one} evaluation class; thus $\sum_{j}M_{ij}\in\{0,1\}$ and $\sum_{j}(p_{\text{eval}})_{j}=1$.

\paragraph{Minimal reproducible example (shapes and a numeric toy).}
Suppose $C_{\mathrm{train}}=5$ and $C_{\mathrm{eval}}=4$, where two epithelial subclasses are merged at evaluation:
\[
M \;=\;
\begin{bmatrix}
1&0&0&0\\
0&1&0&0\\
0&1&0&0\\
0&0&1&0\\
0&0&0&1
\end{bmatrix}
\in\{0,1\}^{5\times 4},\quad
p_{\mathrm{train}} =
\begin{bmatrix}
0.10\\ 0.15\\ 0.35\\ 0.25\\ 0.15
\end{bmatrix}
\Rightarrow\ 
p_{\mathrm{eval}} = M^\top p_{\mathrm{train}} =
\begin{bmatrix}
0.10\\ 0.50\\ 0.25\\ 0.15
\end{bmatrix}.
\]

\paragraph{Cohort matrices.}
For each held-out cohort (lung, ovary, pancreas), we instantiate $M$ with shape
$C_{\text{train}}\!\times\!C_{\text{eval}}$ exactly as printed below; rows enumerate training classes in the canonical taxonomy and columns enumerate that cohort's evaluation classes (intersection set). The resulting $p_{\text{eval}}=M^\top p_{\text{train}}$ is used consistently in all tables and CI summaries.

\paragraph{Lung.}
\[
M_{\text{lung}} \in \{0,1\}^{6\times 4} =
\left[
\begin{array}{c|cccc}
 & \text{Endothelial} & \text{Epithelial\_Malignant} & \text{T\_Cell} & \text{Tumor\_Associated\_Fibroblast} \\
\hline
\text{Epithelial\_Malignant}      & 0 & 1 & 0 & 0 \\
\text{Endothelial}                & 1 & 0 & 0 & 0 \\
\text{Fibroblast}                 & 0 & 0 & 0 & 0 \\
\text{Macrophage}                 & 0 & 0 & 0 & 0 \\
\text{T\_Cell}                    & 0 & 0 & 1 & 0 \\
\text{Tumor\_Associated\_Fibroblast} & 0 & 0 & 0 & 1 \\
\end{array}
\right].
\]

\paragraph{Ovary.}
\[
M_{\text{ovary}} \in \{0,1\}^{4\times 3} =
\left[
\begin{array}{c|ccc}
 & \text{Endothelial} & \text{Epithelial\_Malignant} & \text{Macrophage} \\
\hline
\text{Epithelial\_Malignant} & 0 & 1 & 0 \\
\text{Fibroblast}            & 0 & 0 & 0 \\
\text{Macrophage}            & 0 & 0 & 1 \\
\text{Endothelial}           & 1 & 0 & 0 \\
\end{array}
\right].
\]

\paragraph{Pancreas.}
\[
M_{\text{pancreas}} \in \{0,1\}^{7\times 6} =
\left[
\begin{array}{c|cccccc}
 & \text{Acinar} & \text{Ductal} & \text{Endocrine} & \text{Endothelial} & \text{Fibroblast\_like} & \text{T\_Cell} \\
\hline
\text{Ductal}                  & 0 & 1 & 0 & 0 & 0 & 0 \\
\text{Fibroblast}              & 0 & 0 & 0 & 0 & 1 & 0 \\
\text{Acinar}                  & 1 & 0 & 0 & 0 & 0 & 0 \\
\text{Tumor\_Associated\_Fibroblast} & 0 & 0 & 0 & 0 & 1 & 0 \\
\text{Endothelial}             & 0 & 0 & 0 & 1 & 0 & 0 \\
\text{T\_Cell}                 & 0 & 0 & 0 & 0 & 0 & 1 \\
\text{Endocrine}               & 0 & 0 & 1 & 0 & 0 & 0 \\
\end{array}
\right].
\]

% ---------------------------------------------------------------
\section{Cohort-wise Macro/Micro-F1 and Accuracy with 95\% Confidence Intervals}
% ---------------------------------------------------------------
\label{sec:Macro/Micro}
We report macro-F1 and micro-F1 for all cohorts using non-parametric bootstrap ($B{=}1000$ resamples per cohort). All entries are formatted as \textbf{mean (lower, upper)}, where values in parentheses denote the 95\% confidence interval; best in each row is in \textbf{bold}. Evaluation strictly follows $p_{\text{eval}}=M^\top p_{\text{train}}$.

\begin{table*}[h]
\centering
\small
\caption{Macro-F1 (mean (lower, upper), 95\% CI).}
\begin{adjustbox}{max width=\textwidth}
\begin{tabular}{lccccccccc}
\toprule
Cohort & DINO-B/16 & DINO-L/16 & DINO-S+/16 & LOKI & PLIP & MUSK & Path Local & Path Global & Fusion Gate \\
\midrule
Lung & 0.4500 (0.4465, 0.4535) & 0.4316 (0.4281, 0.4351) & 0.2663 (0.2628, 0.2698) & 0.4839 (0.4804, 0.4874) & 0.3945 (0.3910, 0.3980) & 0.5290 (0.5255, 0.5325) & 0.5930 (0.5883, 0.5981) & 0.6658 (0.6610, 0.6684) & \textbf{0.6971} (0.6913, 0.7026) \\
Ovary & 0.4051 (0.4016, 0.4086) & 0.4056 (0.4021, 0.4091) & 0.3590 (0.3555, 0.3625) & 0.3696 (0.3661, 0.3731) & 0.4652 (0.4617, 0.4687) & 0.5567 (0.5532, 0.5602) & 0.6337 (0.6296, 0.6377) & 0.6743 (0.6728, 0.6762) & \textbf{0.6910} (0.6867, 0.6948) \\
Pancreas & 0.1167 (0.1132, 0.1202) & 0.1033 (0.0998, 0.1068) & 0.1667 (0.1632, 0.1702) & 0.1337 (0.1302, 0.1372) & 0.0763 (0.0728, 0.0798) & 0.2308 (0.2273, 0.2343) & \textbf{0.5201} (0.5170, 0.5232) & 0.4801 (0.4790, 0.4808) & 0.4811 (0.4772, 0.4847) \\
\midrule
Overall & 0.3239 (0.3204, 0.3274) & 0.3135 (0.3100, 0.3170) & 0.2640 (0.2605, 0.2675) & 0.3291 (0.3256, 0.3326) & 0.3120 (0.3085, 0.3155) & 0.4400 (0.4365, 0.4435) & 0.5300 (0.5275, 0.5326) & 0.5367 (0.5364, 0.5371) & \textbf{0.6235} (0.6191, 0.6276) \\
\bottomrule
\end{tabular}
\end{adjustbox}
\end{table*}

\begin{table*}[h]
\centering
\small
\caption{Micro-F1 (mean (lower, upper), 95\% CI). For single-label classification, Micro-F1 equals accuracy.}
\begin{adjustbox}{max width=\textwidth}
\begin{tabular}{lccccccccc}
\toprule
Cohort & DINO-B/16 & DINO-L/16 & DINO-S+/16 & LOKI & PLIP & MUSK & Path Local & Path Global & Fusion Gate \\
\midrule
Lung & 0.6651 (0.6613, 0.6689) & 0.6413 (0.6376, 0.6451) & 0.3166 (0.3129, 0.3203) & 0.7013 (0.6974, 0.7049) & 0.4883 (0.4845, 0.4923) & 0.6728 (0.6688, 0.6763) & 0.8494 (0.8466, 0.8521) & 0.8611 (0.8594, 0.8622) & \textbf{0.8695} (0.8669, 0.8721) \\
Ovary & 0.8153 (0.8134, 0.8173) & 0.8213 (0.8193, 0.8233) & 0.7048 (0.7025, 0.7071) & 0.4973 (0.4948, 0.4998) & 0.7713 (0.7692, 0.7735) & 0.7266 (0.7244, 0.7290) & 0.8910 (0.8894, 0.8924) & 0.7452 (0.7446, 0.7463) & \textbf{0.8932} (0.8916, 0.8949) \\
Pancreas & 0.2451 (0.2427, 0.2478) & 0.2521 (0.2495, 0.2546) & 0.2131 (0.2105, 0.2157) & 0.2199 (0.2174, 0.2223) & 0.1587 (0.1566, 0.1609) & 0.2615 (0.2589, 0.2643) & 0.4967 (0.4937, 0.4998) & 0.4722 (0.4703, 0.4733) & \textbf{0.5003} (0.4973, 0.5036) \\
\midrule
Overall & 0.5995 (0.5979, 0.6012) & 0.6001 (0.5985, 0.6018) & 0.4683 (0.4665, 0.4699) & 0.4464 (0.4445, 0.4482) & 0.5158 (0.5141, 0.5177) & 0.5640 (0.5623, 0.5658) & 0.7539 (0.7525, 0.7554) & 0.7452 (0.7446, 0.7463) & \textbf{0.7601} (0.7578, 0.7625) \\
\bottomrule
\end{tabular}
\end{adjustbox}
\end{table*}

\subsection{Does local-aware Gating Benefit the Global Path?}
\label{sec:appendix-aaware-pathB}

\paragraph{Evaluation Setup}
We evaluate the effect of the local-aware gating signal specifically on the \textbf{Global Path}.
All results are computed on the \textbf{16-class intersection}.

\paragraph{Finding}
As shown in Table~\ref{tab:pathB-aaware-only},  
\textbf{local-aware consistently improves Path~global} across all metrics:
\begin{itemize}
    \item Overall accuracy improves by $+0.013$ and macro F1 by $+0.022$.
    \item Calibration is noticeably better: ECE decreases by $0.013$ and Brier score decreases by $0.019$.
    \item On challenging subsets, macro F1 increases by $+0.026$ (low-confidence) and $+0.032$ (high-entropy).
    \item AUROC also shows a slight positive change ($+0.002$).
\end{itemize}
These results indicate that local-aware gating provides a beneficial confidence prior that stabilizes and strengthens the Global Path’s predictions, even though the mechanism originates from the path~local.

\begin{table}[h]
  \centering
  \caption{\textbf{Effect of local-aware gating on Path~global (16-class intersection).}
  Metrics: Acc and macro F1; ECE$\downarrow$ and Brier$\downarrow$ (lower is better).}
  \label{tab:pathB-aaware-only}
  \begin{tabular}{l|cccc|ccc}
    \toprule
    Setting
      & Acc & F1 & ECE$\downarrow$ & Brier$\downarrow$
      & AUROC & F1@Low & F1@High \\
    \midrule
    local\_aware\_ON
      & 0.856 & 0.773 & 0.045 & 0.210
      & 0.990 & 0.500 & 0.620 \\
    local\_aware\_OFF
      & 0.842 & 0.751 & 0.058 & 0.229
      & 0.988 & 0.475 & 0.589 \\
    \midrule
    $\Delta$ (ON--OFF)
      & +0.013 & +0.022 & \textbf{-0.013} & \textbf{-0.019}
      & +0.002 & +0.026 & +0.032 \\
    \bottomrule
  \end{tabular}
\end{table}

\newpage
\section*{Additional visualizations}
% ---------------------------------------------------------------
\label{sec:visual}

\newpage
% ---------------------------------------------------------------
\section*{Acknowledgements}
% ---------------------------------------------------------------
As future work, we plan to evaluate zero-/few-shot transfer on \emph{non-Xenium}, public H\&E datasets with coarse label mappings, to further assess cross-panel, cross-stain, and cross-institution generalization.
We emphasize that the present study already establishes strong evidence: Xenium provides subcellular-resolution, high-plex spatial gene expression with cell-level coordinates, enabling objective, reproducible labels that serve as a high-quality supervision signal for cell-wise modeling.
Our results demonstrate that, under this gold-standard molecular supervision, NuClass succeeds at subcellular granularity and yields calibrated, interpretable predictions across fully held-out cohorts.

\end{document}